# Policy Iteration for Decentralized Control of Markov Decision Processes

**Daniel S. Bernstein**                                      BERN@CS.UMASS.EDU

**Christopher Amato**                                        CAMATO@CS.UMASS.EDU
*Department of Computer Science*
*University of Massachusetts*
*Amherst, MA 01003 USA*

**Eric A. Hansen**                                           HANSEN@CSE.MSSTATE.EDU
*Department of CS and Engineering*
*Mississippi State University*
*Mississippi State, MS 39762 USA*

**Shlomo Zilberstein**                                       SHLOMO@CS.UMASS.EDU
*Department of Computer Science*
*University of Massachusetts*
*Amherst, MA 01003 USA*

## Abstract

Coordination of distributed agents is required for problems arising in many areas, including multi-robot systems, networking and e-commerce. As a formal framework for such problems, we use the decentralized partially observable Markov decision process (DEC-POMDP). Though much work has been done on optimal dynamic programming algorithms for the single-agent version of the problem, optimal algorithms for the multiagent case have been elusive. The main contribution of this paper is an optimal policy iteration algorithm for solving DEC-POMDPs. The algorithm uses stochastic finite-state controllers to represent policies. The solution can include a correlation device, which allows agents to correlate their actions without communicating. This approach alternates between expanding the controller and performing value-preserving transformations, which modify the controller without sacrificing value. We present two efficient value-preserving transformations: one can reduce the size of the controller and the other can improve its value while keeping the size fixed. Empirical results demonstrate the usefulness of value-preserving transformations in increasing value while keeping controller size to a minimum. To broaden the applicability of the approach, we also present a heuristic version of the policy iteration algorithm, which sacrifices convergence to optimality. This algorithm further reduces the size of the controllers at each step by assuming that probability distributions over the other agents' actions are known. While this assumption may not hold in general, it helps produce higher quality solutions in our test problems.

## 1. Introduction

Markov decision processes (MDPs) provide a useful framework for solving problems of sequential decision making under uncertainty. In some settings, agents must base their decisions on partial information about the system state. In that case, it is often better to use the more general framework of partially observable Markov decision processes (POMDPs). Even more general are problems in which a team of decision makers, each with its own local observations, must act together. Domains in which these types of problems arise include networking, multi-robot coordination, e-commerce, and space exploration systems. The decentralized partially observable Markov decision process (DEC-POMDP) provides an effective framework to model such problems. Though this model has been recognized for decades (Witsenhausen, 1971), there has been little work on provably optimal algorithms for it.

On the other hand, POMDPs have been studied extensively over the past few decades (Smallwood & Sondik, 1973; Simmons & Koenig, 1995; Cassandra, Littman, & Zhang, 1997; Hansen, 1998a; Bonet & Geffner, 2000; Poupart & Boutilier, 2003; Feng & Zilberstein, 2004; Smith & Simmons, 2005; Smith, Thompson, & Wettergreen, 2007). It is well known that a POMDP can be reformulated as an equivalent *belief-state* MDP. A belief-state MDP cannot be solved in a straightforward way using MDP methods because it has a continuous state space. However, Smallwood and Sondik showed how to implement value iteration by exploiting the piecewise linearity and convexity of the value function. This work opened the door for many algorithms, including approximate approaches and policy iteration algorithms in which the policy is represented using a finite-state controller.

Extending dynamic programming for POMDPs to the multiagent case is not straightforward. For one thing, it is not clear how to define a belief state and consequently form a belief-state MDP. With multiple agents, each agent has uncertainty about the observations and beliefs of the other agents. Furthermore, the finite-horizon DEC-POMDP problem with just two agents is complete for a higher complexity class than the single-agent version (Bernstein, Givan, Immerman, & Zilberstein, 2002), indicating that these are fundamentally different problems.

In this paper, we describe an extension of the policy iteration algorithm for single agent POMDPs to the multiagent case. As in the single agent case, our algorithm converges in the limit, and thus serves as the first nontrivial optimal algorithm for infinite-horizon DEC-POMDPs. A few optimal approaches (Hansen, Bernstein, & Zilberstein, 2004; Szer, Charpillet, & Zilberstein, 2005) and several approximate algorithms have been developed for finite-horizon DEC-POMDPs (Peshkin, Kim, Meuleau, & Kaelbling, 2000; Nair, Pynadath, Yokoo, Tambe, & Marsella, 2003; Emery-Montemerlo, Gordon, Schnieder, & Thrun, 2004; Seuken & Zilberstein, 2007), but only locally optimal algorithms have been proposed for the infinite-horizon case (Bernstein, Hansen, & Zilberstein, 2005; Szer & Charpillet, 2005; Amato, Bernstein, & Zilberstein, 2007).

In our algorithmic framework, policies are represented using stochastic finite-state controllers. A simple way to implement this is to give each agent its own local controller. In this case, the agents' policies are all independent. A more general class of policies includes those which allow agents to share a common source of randomness without sharing observations. We define this class formally, using a shared source of randomness called a *correlation device*. The use of correlated stochastic policies in the DEC-POMDP context is novel. The importance of correlation has been recognized in the game theory community (Aumann, 1974), but there has been little work on algorithms for finding correlated policies.

Each iteration of the algorithm consists of two phases. These are *exhaustive backups*, which add nodes to the controller, and *value-preserving transformations*, which change the controller without sacrificing value. We first provide a novel exposition of existing single-agent algorithms using this two-phase view, and then we go on to describe the multiagent extension.

There are many possibilities for value-preserving transformations. In this paper, we describe two different types, both of which can be performed efficiently using linear programming. The first type allows us to remove nodes from the controller, and the second allows us to improve the value of the controller while keeping its size fixed. Our empirical results demonstrate the usefulness of value-preserving transformations in obtaining high values while keeping controller size to a minimum.

We note that this work serves to unify and generalize previous work on dynamic programming for DEC-POMDPs. The first algorithm for the finite-horizon case (Hansen et al., 2004) can be extended to the infinite-horizon case and viewed as interleaving exhaustive backups and controller reductions. The *bounded policy iteration* algorithm for DEC-POMDPs (Bernstein et al., 2005), which extends a POMDP algorithm proposed by Poupart and Boutilier (2003), can be viewed through the lens of our framework as repeated application of a specific value-preserving transformation.

Because the optimal algorithm will not usually be able to return an optimal solution in practice, we also introduce a heuristic version of the policy iteration algorithm. This approach makes use of initial state information to focus policy search and further reduces controller size at each step. To accomplish this, a forward search from the initial state distribution is used to construct a set of belief points an agent would visit assuming the other agents use given fixed policies. This search is conducted for each agent and then policy iteration takes place while using the belief points to guide the removal of controller nodes. The assumption that other agents use fixed policies causes the algorithm to no longer be optimal, but it performs well in practice. We show that more concise and higher-valued solutions can be produced compared to the optimal method before resources are exhausted.

The remainder of the paper is organized as follows. Section 2 introduces the formal models of sequential decision making. Section 3 contains a novel presentation of existing dynamic programming algorithms for POMDPs. In section 4, we present an extension of policy iteration for POMDPs to the DEC-POMDP case, along with a convergence proof. We discuss the heuristic version of policy iteration in section 5, followed by experiments using policy iteration and heuristic policy iteration in section 6. Finally, section 7 contains the conclusion and a discussion of possible future work.

## 2. Formal Model of Distributed Decision Making

We begin with a description of the formal framework upon which our work is based. This framework extends the well-known Markov decision process to allow for distributed policy execution. We also define an optimal solution for this model and discuss two different representations for these solutions.

### 2.1 Decentralized POMDPs

A *decentralized partially observable Markov decision process (DEC-POMDP)* is defined formally as a tuple $\langle I, S, \vec{A}, T, R, \vec{\Omega}, O \rangle$, where

- $I$ is a finite set of agents.

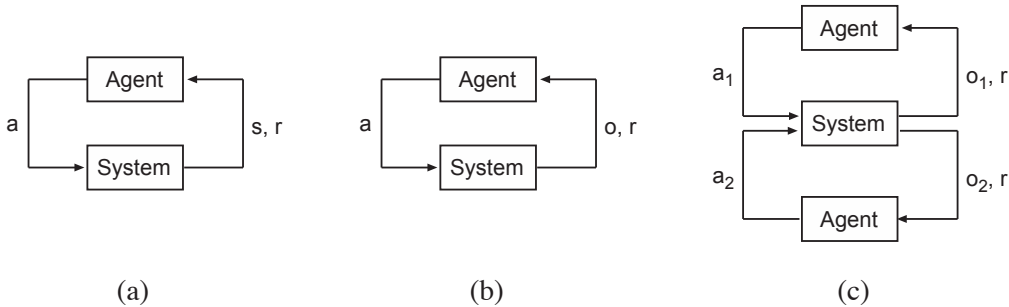

Figure 1: (a) Markov decision process. (b) Partially observable Markov decision process.
(c) Decentralized partially observable Markov decision process with two agents.

- $S$ is a finite set of states, with distinguished initial state $s_0$.

- $\vec{A} = \times_{i \in I} A_i$ is a set of joint actions, where $A_i$ is the set of actions for agent $i$.

- $T : S \times \vec{A} \to \Delta S$ is the state transition function, defining the distributions of states that result from starting in a given state and each agent performing an action.

- $R : S \times \vec{A} \to \Re$ is the reward function for the set of agents for each set of joint actions and each state.

- $\vec{\Omega} = \times_{i \in I} \Omega_i$ is a set of joint observations, where $\Omega_i$ contains observations for agent $i$.

- $O : \vec{A} \times S \to \Delta \vec{\Omega}$ is an observation function, defining the distributions of observations for the set of agents that result from each agent performing an action and ending in a given state.

The special case of a DEC-POMDP in which there is only one agent is called a *partially observable Markov decision process (POMDP)*.

In this paper, we consider the case in which the process unfolds over an infinite sequence of stages. At each stage, all agents simultaneously select an action, and each receives the global reward based on the reward function and a local observation based on the observation function. Thus, the transitions, rewards and observations depend on the actions of all agents, but each agent must act based only on local observations. This is illustrated in Figure 1. The objective of the agents is to maximize the expected discounted sum of rewards that are received, thus it is a cooperative framework. We denote the discount factor $\beta$ and require that $0 \le \beta < 1$.

In a DEC-POMDP, the decisions of each agent affect all the agents in the domain, but due to the decentralized nature of the model each agent must choose actions based solely on local information. Because each agent receives a separate observation that does not usually provide sufficient information to efficiently reason about the other agents, solving a DEC-POMDP optimally becomes very difficult. For example, each agent may receive a different piece of information that does not allow a common state estimate or any estimate of the other agents' decisions to be calculated. These single estimates are crucial in single agent problems, as they allow the agent's history to be summarize concisely, but they are not generally available in DEC-POMDPs. This is seen in the complexity of the finite-horizon problem with at least two agents, which is NEXP-complete (Bernstein et al., 2002) and thus in practice may require double exponential time. Like the infinite-horizon POMDP, optimally solving an infinite-horizon DEC-POMDP is undecidable as it may require infinite resources, but our method is able to provide a solution within $\epsilon$ of the optimal with finite time and memory. Nevertheless, introducing multiple decentralized agents causes a DEC-POMDP to be significantly more difficult than a single agent POMDP.

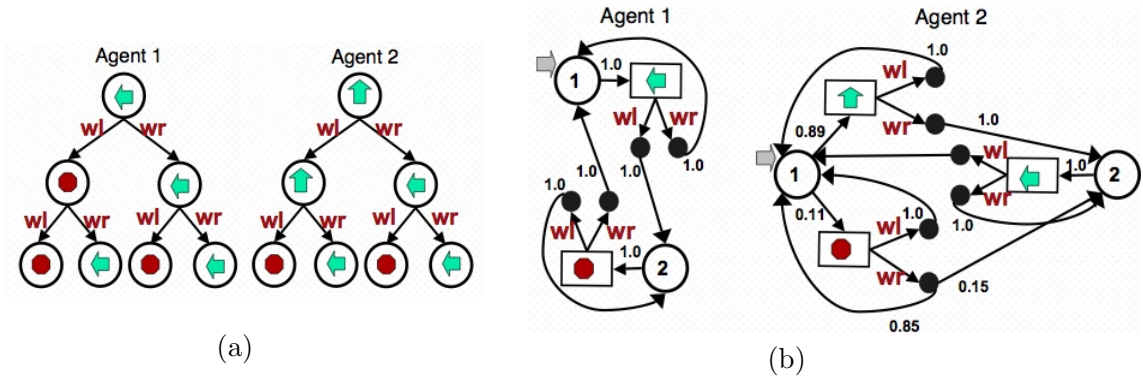

(a)  (b)

Figure 2: A set of horizon three policy trees (a) and two node stochastic controllers (b) for a two agent DEC-POMDP.

## 2.2 Solution Representations

A *local policy* for an agent is a mapping from local action and observation histories to actions while a *joint policy* is a set of policies, one for each agent in the problem. As mentioned above, an optimal solution for a DEC-POMDP is the joint policy that maximizes the expected sum of rewards that are received over the finite or infinite steps of the problem. In infinite-horizon problems, the rewards are discounted to maintain a finite sum. Thus, an optimal solution is a joint policy that provides the highest value starting at the given initial state of the problem.

For finite-horizon problems, local policies can be represented using a policy tree as seen in Figure 2a. Actions are represented by the arrows or stop figures (where each agent can move in the given direction or stay where it is) and observations are labeled "wl" and "wr" for seeing a wall on the left or the right respectively. Using this representation, an agent takes the action defined at the root node and then after seeing an observation, chooses the next action that is defined by the respective branch. This continues until the action at a leaf node is executed. For example, agent 1 would first move left and then if a wall is seen on the right, the agent would move left again. If a wall is now seen on the left, the agent does not move on the final step. A policy tree is a record of the the entire local history for an agent up to some fixed horizon and because each tree is independent of the others it can be executed in a decentralized manner. While this representation is useful for finite-horizon problems, infinite-horizon problems would require trees of infinite height.

Another option used in this paper is to condition action selection on some internal memory state. These solutions can be represented as a set of local finite-state controllers (seen in Figure 2b). The controllers operate in a very similar way to the policy trees in that there is a designated initial node and following the action selection at that node, the controller transitions to the next node depending on the observation seen. This continues for the infinite steps of the problem. Throughout this paper, controller states will be referred to as nodes to help distinguish them from system states.

An infinite number of nodes may be required to define an optimal infinite-horizon DEC-POMDP policy, but we will discuss a way to produce solutions within $\epsilon$ of the optimal with a fixed number of nodes. While deterministic action selection and node transitions are sufficient to define this $\epsilon$-optimal policy, when memory is limited stochastic action selection and node transition may be beneficial. A simple example illustrating this for POMDPs is given by Singh (1994), which can be easily extended to DEC-POMDPs. Intuitively, randomness can help an agent to break out of costly loops that result from forgetfulness.

A formal description of stochastic controllers for POMDPs and DEC-POMDPs is given in sections 3.2.1 and 4.1.1 respectively, but an example can be seen in Figure 2b. Agent 2 begins at node 1 and moves up with probability 0.89 and stays in place with probability 0.11. If the agent stayed in place and a wall was then seen on the left (observation "wl"), on the next step, the controller would transition to node 1 and the agent would use the same distribution of actions again. If a wall was seen on the right instead (observation "wr"), there is a 0.85 probability that the controller will transition back to node 1 and a 0.15 probability that the controller will transition to node 2 for the next step. The finite-state controller allows an infinite-horizon policy to be represented compactly by remembering some aspects of the agent's history without representing the entire local history.

## 3. Centralized Dynamic Programming

In this section, we cover the main concepts involved in dynamic programming for the single agent case. This will provide a foundation for the multiagent dynamic programming algorithm described in the following section.

### 3.1 Value Iteration for POMDPs

Value iteration can be used to solve POMDPs optimally. This algorithm is more complicated than its MDP counterpart, and does not have efficiency guarantees. However, in practice it can provide significant leverage in solving POMDPs.

We begin by explaining how every POMDP has an equivalent MDP with a continuous state space. Next, we describe how the value functions for this MDP have special structure that can be exploited. These ideas are central to the value iteration algorithm.

#### 3.1.1 Belief State MDPs

A convenient way to summarize the observation history of an agent in a POMDP is through a *belief state*, which is a distribution over system states. As it receives observations, the agent can update its belief state and then remove its observations from memory. Let $b$ denote a belief state, and let $b(s)$ represent the probability assigned to state $s$ by $b$. If an agent chooses action $a$ from belief state $b$ and subsequently observes $o$, each component of the successor belief state obeys the equation

$$b'(s') = \frac{P(o|a, s') \sum_{s \in S} P(s'|s, a) b(s)}{P(o|b, a)},$$

where

$$P(o|b, a) = \sum_{s' \in S} \left[ P(o|a, s') \sum_{s \in S} P(s'|s, a) b(s) \right].$$

Note that this is a simple application of Bayes' rule.

It was shown by Astrom (1965) that a belief state constitutes a sufficient statistic for the agent's observation history, and it is possible to define an MDP over belief states as follows. A *belief-state MDP* is a tuple $\langle \Pi, A, T, R \rangle$, where

- $\Pi$ is the set of distributions over $S$.

- $A$ is the set of actions (same as before).

- $T(b, a, b')$ is the transition function, defined as

$$T(b, a, b') = \sum_{o \in O} P(b'|b, a, o) P(o|b, a).$$

- $R(b, a)$ is a reward function, defined as

$$R(b, a) = \sum_{s \in S} b(s) R(s, a).$$

When combined with belief-state updating, an optimal solution to this MDP can be used as an optimal solution to the POMDP from which it was constructed. However, since the belief state MDP has a continuous, $|S|$-dimensional state space, traditional MDP techniques are not immediately applicable.

Fortunately, dynamic programming can be used to find a solution to the belief state MDP. The key result in making dynamic programming practical was proved by Smallwood and Sondik (1973), who showed that the Bellman operator preserves piecewise linearity and convexity of a value function. Starting with a piecewise linear and convex representation of $V^t$, the value function $V^{t+1}$ is piecewise linear and convex, and can be computed in finite time.

To represent a piecewise linear and convex value function, one need only store the value of each facet for each system state. Denoting the set of facets $\Gamma$, we can store $|\Gamma|$ $|S|$-dimensional vectors of real values. For any single vector, $\gamma \in \Gamma$, we can define its value at the belief state $b$ with $V(b, \gamma) = \sum_{s \in S} b(s) \gamma(s)$. Thus, to go from a set of vectors to the value of a belief state, we use the equation

$$V(b) = \max_{\gamma} \sum_{s \in S} b(s) \gamma(s).$$

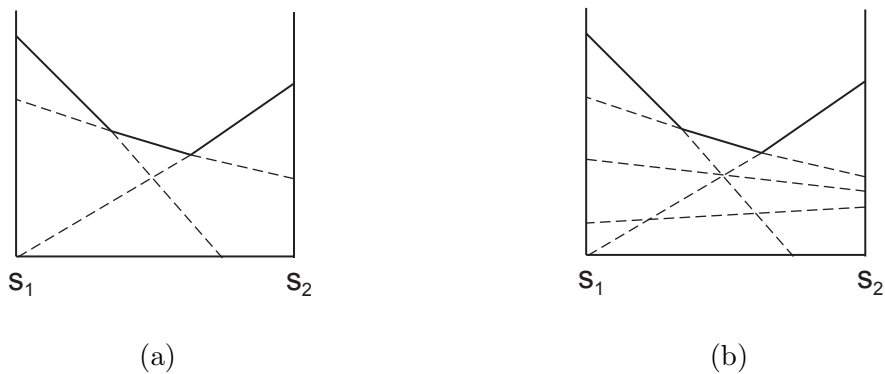

(a)                                                     (b)

Figure 3: A piecewise linear and convex value function for a POMDP with two states (a) and a non-minimal representation of a piecewise linear and convex value function for a POMDP (b).

Figure 3a shows a piecewise linear and convex value function for a POMDP with two states.

Smallwood and Sondik proved that the optimal value function for a finite-horizon POMDP is piecewise linear and convex. The optimal value function for an infinite-horizon POMDP is convex, but may not be piecewise linear. However, it can be approximated arbitrarily closely by a piecewise linear and convex value function, and the value iteration algorithm constructs closer and closer approximations, as we shall see.

### 3.1.2 PRUNING VECTORS

Every piecewise linear and convex value function has a minimal set of vectors $\Gamma$ that represents it. Of course, it is possible to use a non-minimal set to represent the same function. This is illustrated in Figure 3b. Note that the removal of certain vectors does not change the value of any belief state. Vectors such as these are not necessary to keep in memory. Formally, we say that a vector $\gamma$ is *dominated* if for all belief states $b$, there is a vector $\hat{\gamma} \in \Gamma \setminus \gamma$ such that $V(b, \gamma) \leq V(b, \hat{\gamma})$.

Because dominated vectors are not necessary, it would be useful to have a method for removing them. This task is often called *pruning*, and has an efficient algorithm based on linear programming. For a given vector $\gamma$, the linear program in Table 1 determines whether $\gamma$ is dominated. If variables can be found to make $\epsilon$ positive, then adding $\gamma$ to the set improves the value function at some belief state. If not, then $\gamma$ is dominated.

This gives rise to a simple algorithm for pruning a set of vectors $\tilde{\Gamma}$ to obtain a minimal set $\Gamma$. The algorithm loops through $\tilde{\Gamma}$, removes each vector $\gamma \in \tilde{\Gamma}$, and solves the linear program using $\gamma$ and $\tilde{\Gamma} \setminus \gamma$. If $\gamma$ is not dominated, then it is returned to $\tilde{\Gamma}$.

It turns out that there is an equivalent way to characterize dominance that can be useful. Recall that for a vector to be dominated, there does not have to be a single vector that has value at least as high for all states. It is sufficient for there to exist a set of vectors such that for all belief states, one of the vectors in the set has value at least as high as the vector in question.

Variables: $\epsilon$, $b(s)$

Objective: Maximize $\epsilon$.

Improvement constraints:

$$\forall \hat{\gamma} \quad \sum_s b(s)\hat{\gamma}(s) + \epsilon \leq \sum_s b(s)\gamma(s)$$

Probability constraints:

$$\sum_s b(s) = 1, \quad \forall s \quad b(s) \geq 0$$

Table 1: The linear program for testing whether a vector $\gamma$ is dominated.

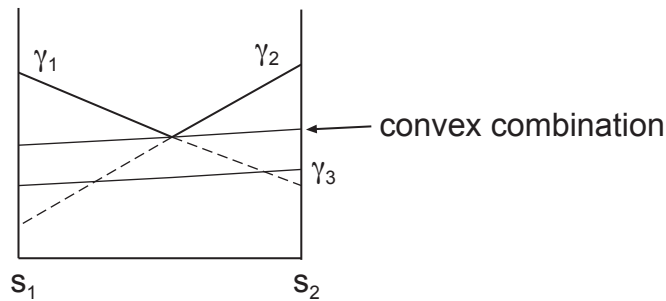

Figure 4: The dual interpretation of dominance. Vector $\gamma_3$ is dominated at all belief states by either $\gamma_1$ or $\gamma_2$. This is equivalent to the existence of a convex combination of $\gamma_1$ and $\gamma_2$ which dominates $\gamma_3$ for all belief states.

It can be shown that such a set exists if and only if there is some convex combination of vectors that has value at least as high as the vector in question for all states. This is shown graphically in Figure 4. If we take the dual of the linear program for dominance given in the previous section, we get a linear program for which the solution is a vector of probabilities for the convex combination. This dual view of dominance was first used in a POMDP context by Poupart and Boutilier (2003), and is useful for policy iteration, as will be explained later.

### 3.1.3 Dynamic Programming Update

In this section, we describe how to implement a *dynamic programming update* to go from a value function $V_t$ to a value function $V_{t+1}$. In terms of implementation, our aim is to take a minimal set of vectors $\Gamma_t$ that represents $V_t$ and produce a minimal set of vectors $\Gamma_{t+1}$ that represents $V_{t+1}$.

Each vector that could potentially be included in $\Gamma_{t+1}$ represents the value of an action $a$ and assignment of vectors in $\Gamma_t$ to observations. A combination of an action and transition rule will hereafter be called a *one-step policy*. The value vector for a one-step policy can be determined by considering the action taken, the resulting state transitioned to and observation seen and the value of the assigned vector at step $t$. This is given via the equation

$$\gamma_i^{t+1}(s) = R(s, \alpha(i)) + \beta \sum_{s',o} P(s'|s, \alpha(i)) P(o|\alpha(i), s') \gamma_{\tau(i,o)}^t(s'),$$

where $i$ is the index of the vector, $\alpha(i)$ is its action, and $\tau(i, o)$ is the index of the vector in $\Gamma_t$ to which to transition upon receiving observation $o$ and $\beta$ is the discount factor. More details on the derivation and use of this formula are provided by Zhang and Zhang (2001).

There are $|A||\Gamma_t|^{|\Omega|}$ possible one-step policies. A simple way to construct $\Gamma_{t+1}$ is to evaluate all possible one-step policies and then apply a pruning algorithm such as Lark's method (Lark III, 1990). Evaluating the entire set of one-step policies will hereafter be called performing an *exhaustive backup*. It turns out that there are ways to perform a dynamic programming update without first performing an exhaustive backup. Below we describe two approaches to doing this.

The first approach uses the fact that it is simple to find the optimal vector for any particular belief state. For a belief state $b$, an optimal action can be determined via the equation

$$\alpha = \text{argmax}_{a \in A} \left[ R(b, a) + \beta \sum_{o \in \Omega} P(o|b, a) V^t(T(b|a, o)) \right].$$

For each observation $o$, there is a subsequent belief state, which can be computed using Bayes' rule. To get an optimal transition rule, $\tau(o)$, we take the optimal vector for the belief state corresponding to $o$.

Since the backed-up value function has finitely many vectors, there must be a finite set of belief states for which backups must be performed. Algorithms which identify these belief states include Smallwood and Sondik's "one-pass" algorithm (1973), Cheng's linear support and relaxed region algorithms (Cheng, 1988), and Kaelbling, Cassandra and Littman's Witness algorithm (1998).

The second approach is based on generating and pruning sets of vectors. Instead of generating all vectors and then pruning, these techniques attempt to prune during the generation phase. The first algorithm along these lines was the incremental pruning algorithm (Cassandra et al., 1997). Recently, improvements have been made to this approach (Zhang & Lee, 1998; Feng & Zilberstein, 2004, 2005).

It should be noted that there are theoretical complexity barriers for DP updates. Littman et al. (1995) showed that under certain widely believed complexity theoretic assumptions, there is no algorithm for performing a DP update that is worst-case polynomial in all the quantities involved. Despite this fact, dynamic programming updates have been successfully implemented as part of the value iteration and policy iteration algorithms, which will be described in the subsequent sections.

### 3.1.4 Value Iteration

To implement value iteration, we simply start with an arbitrary piecewise linear and convex value function, and proceed to perform DP updates. This corresponds to value iteration in the equivalent belief state MDP, and thus converges to an $\epsilon$-optimal value function after a finite number of iterations.

Value iteration returns a value function, but a policy is needed for execution. As in the MDP case, we can use one-step lookahead, using the equation

$$\delta(b) = \text{argmax}_{a \in A} \left[ \sum_{s \in S} R(s, a)b(s) + \beta \sum_{o \in \Omega} P(o|b, a)V(\tau(b, o, a)) \right],$$

where $\tau(b, o, a)$ is the belief state resulting from starting in belief state $b$, taking action $a$, and receiving observation $o$. We note that a state estimator must be used as well to track the belief state. Using the fact that each vector corresponds to a one-step policy, we can extract a policy from the value of the vectors:

$$\delta(b) = \alpha \left( \text{argmax}_k \sum_s b(s)\gamma_k(s) \right)$$

While the size of the resulting set of dominant vectors may remain exponential, in many cases it is much smaller. This can significantly simplify computation.

As in the completely observable case, the Bellman residual provides a bound on the distance to optimality. Recall that the Bellman residual is the maximum distance across all belief states between the value functions of successive iterations. It is possible to find the maximum distance between two piecewise linear and convex functions in polynomial time with an algorithm that uses linear programming (Littman et al., 1995).

## 3.2 Policy Iteration for POMDPs

With value iteration, a POMDP is viewed as a belief-state MDP, and a policy is a mapping from belief states to actions. An early policy iteration algorithm developed by Sondik used this policy representation (Sondik, 1978), but it was very complicated and did not meet with success in practice. We shall describe a different approach that has performed better on test problems. With this approach, a policy is represented as a finite-state controller.

### 3.2.1 Finite-State Controllers

Using a finite-state controller, an agent has a finite number of internal states. Its actions are based only on its internal state, and transitions between internal states occur when observations are received. Internal states provide agents with a kind of memory, which can be crucial for difficult POMDPs. Of course, an agent's memory is limited by the number of internal states it possesses. In general, an agent cannot remember its entire history of observations, as this would require infinitely many internal states. An example of a finite-state controller can be seen by considering only one agent's controller in Figure 2b. The operation of a single controller is the same as that for each agent in the decentralized case.

We formally define a *controller* as a tuple $\langle Q, \Omega, A, \psi, \eta \rangle$, where

- $Q$ is a finite set of controller nodes.

- $\Omega$ is a set of inputs, taken to be the observations of the POMDP.

- $A$ is a set of outputs, taken to be the actions of the POMDP.

- $\psi : Q \to \Delta A$ is an action selection function, defining the distribution of actions selected at each node.

- $\eta : Q \times A \times \Omega \to \Delta Q$ is a transition function, defining the distribution of resulting nodes for each initial node and action taken.

For each state and starting node of the controller, there is an expected discounted sum of rewards over the infinite horizon. It can be computed using the following system of linear equations, one for each $s \in S$ and $q \in Q$:

$$V(s,q) = \sum_a P(a|q) \left[ R(s,a) + \beta \sum_{s',o,q'} P(o,s'|s,a)P(q'|q,a,o)V(s',q') \right].$$

Where $P(a|q)$ is the probability action $a$ will be taken in node $q$ and $P(q'|q,a,o)$ is the probability the controller will transition to node $q'$ from node $q$ after action $a$ was taken and $o$ was observed.

We sometimes refer to the value of the controller at a belief state. For a belief state $b$, this is defined as

$$V(b) = \max_q \sum_s b(s)V(s,q).$$

Thus, it is assumed that, given an initial state distribution, the controller is started in the node which maximizes value from that distribution. Once execution has begun, however, there is no belief state updating. In fact, it is possible for the agent to encounter the same belief state twice and be in a different internal state each time.

### 3.2.2 Algorithmic Framework

We will describe the policy iteration algorithm in abstract terms, focusing on the key components necessary for convergence. In subsequent sections, we present different possibilities for implementation.

Policy iteration takes as input an arbitrary finite-state controller. The first phase of an iteration consists of evaluating the controller, as described above. Recall that value iteration was initialized with an arbitrary piecewise linear and convex value function, represented by a set of vectors. In policy iteration, the piecewise linear and convex value function arises out of evaluation of the controller. Each controller node has a value when paired with each state. Thus, each node has a corresponding vector and thus a linear value function over belief state space. Choosing the best node for each belief state yields a piecewise linear and convex value function.

The second phase of an iteration is the dynamic programming update. In value iteration, an update produces an improved set of vectors, where each vector corresponds to a deterministic one-step policy. The same set of vectors is produced in this case, but the

---

Input: A finite state controller, and a parameter $\epsilon$.

1. Evaluate the finite-state controller by solving a system of linear equations.

2. Perform a dynamic programming update to add a set of deterministic nodes to the controller.

3. Perform value-preserving transformations on the controller.

4. Calculate the Bellman residual. If it is less than $\epsilon(1 - \beta)/2\beta$, then terminate. Otherwise, go to step 1.

Output: An $\epsilon$-optimal finite-state controller.

---

Table 2: Policy Iteration for POMDPs.

actions and transition rules for the one-step policy cannot be removed from memory. Each new vector is actually a node that gets added to the controller. All of the probability distributions for the added nodes are deterministic. That is, a exhaustive backup in this context creates a new node for each possible action and possible combinations of observations and deterministic transitions into the current controller. This results in the same one-step policies being considered as in the dynamic programming update described above. As there are $|A||\Gamma_t|^{|\Omega|}$ possible one-step polices, this number also defines the number of new nodes added to the controller after an exhaustive backup.

Finally, additional operations are performed on the controller. There are many such operations, and we describe two possibilities in the following section. The only restriction placed on these operations is that they do not decrease the value for any belief state. Such an operation is denoted a *value-preserving transformation*.

The complete algorithm is outlined in Table 2. It is guaranteed to converge to a finite-state controller that is $\epsilon$-optimal for all belief states within a finite number of steps. Furthermore, the Bellman residual can be used to obtain a bound on the distance to optimality, as with value iteration.

### 3.2.3 Controller Reductions

In performing a DP update, potential nodes that are dominated do not get added to the controller. However, after the update is performed, some of the old nodes may have become dominated. These nodes cannot simply be removed, however, as other nodes may transition into them. This is where the dual view of dominance is useful. Recall that if a node is dominated, then there is a convex combination of other nodes with value at least as high from all states. Thus, we can remove the dominated node and *merge* it into the dominating convex combination by changing transition probabilities accordingly. This operation was proposed by Poupart and Boutilier (2003) and built upon earlier work by Hansen (1998b).

Formally, a *controller reduction* attempts to replace a node $q \in Q$ with a distribution $P(\hat{q})$ over nodes $\hat{q} \in Q \setminus q$ such that for all $s \in S$,

$$V(s, q) \leq \sum_{\hat{q} \in Q \setminus q} P(\hat{q})V(s, \hat{q}).$$

Variables: $\epsilon, x(\hat{\gamma})$

Objective: Maximize $\epsilon$

Improvement constraints:

$$\forall s \quad V(s, \gamma) + \epsilon \leq \sum_{\hat{\gamma}} x(\hat{\gamma}) V(s, \hat{\gamma})$$

Probability constraints:

$$\sum_{\hat{\gamma}} x(\hat{\gamma}) = 1, \qquad \forall \hat{\gamma} \quad x(\hat{\gamma}) \geq 0$$

Table 3: The dual linear program for testing dominance for the vector $\gamma$. The variable $x(\hat{\gamma})$ represents $P(\hat{\gamma})$.

This can be achieved by solving the linear program in Table 3. As nodes are used rather than vectors, we replace $x(\hat{\gamma})$ with $x(\hat{q})$ in the dual formulation which provides a probability distribution of nodes which dominate node $q$. Rather than transitioning into $q$, this distribution can then be used instead. It can be shown that if such a distribution is found and used for merging, the resulting controller is a value-preserving transformation of the original one.

### 3.2.4 Bounded Backups

In the previous section, we described a way to reduce the size of a controller without sacrificing value. The method described in this section attempts to increase the value of the controller while keeping its size fixed. It focuses on one node at a time, and attempts to change the parameters of the node such that the value of the controller is at least as high for all belief states. The idea for this approach originated with Platzman (1980), and was made efficient by Poupart and Boutilier (2003).

In this method, a node $q$ is chosen, and parameters for the conditional distribution $P(a, q'|q, o)$ are to be determined. Determining these parameters works as follows. We assume that the original controller will be used from the second step on, and try to replace the parameters for $q$ with better ones for just the first step. In other words, we look for parameters which satisfy the following inequality:

$$V(s, q) \leq \sum_a P(a|q) \left[ R(s, a) + \beta \sum_{s', o, q'} P(q'|q, a, o) P(o, s'|s, a) V(s', q') \right]$$

for all $s \in S$. Note that the inequality is always satisfied by the original parameters. However, it is often possible to get an improvement.

The new parameters can be found by solving a linear program, as shown in Table 4. Note that the size of the linear program is polynomial in the sizes of the POMDP and the controller. We call this process a *bounded backup* because it acts like a dynamic programming

Variables: $\epsilon, x(a), x(a, o, q')$

Objective: Maximize $\epsilon$

Improvement constraints:

$$\forall s \quad V(s, q) + \epsilon \leq \sum_a \left[ x(a) R(s, a) + \beta \sum_{s', o, q'} x(a, o, q') P(o, s' | s, a) V(s', q') \right]$$

Probability constraints:

$$\sum_a x(a) = 1, \quad \forall a, o \quad \sum_{q'} x(a, o, q') = x(a)$$

$$\forall a \quad x(a) \geq 0, \quad \forall a, o, q' \quad x(a, o, q') \geq 0$$

Table 4: The linear program to be solved for a bounded backup. The variable $x(a)$ represents $P(a|q)$, and the variable $x(a, o, q')$ represents $P(a, q'|q, o)$.

backup with memory constraints. To see this, consider the set of nodes generated by a DP backup. These nodes dominate the original nodes across all belief states, so for every original node, there must be a convex combination of the nodes in this set that dominate the original node for all states. A bounded backup finds such a convex combination.

It can be shown that a bounded backup yields a value-preserving transformation. Repeated application of bounded backups can lead to a local optimum, at which none of the nodes can be improved any further. Poupart and Boutilier (2003) showed that a local optimum has been reached when each node's value function is touching the value function produced by performing a full DP backup. This is illustrated in Figure 5.

## 4. Decentralized Dynamic Programming

In the previous section, we presented dynamic programming for POMDPs. A key part of POMDP theory is the fact that every POMDP has an equivalent belief-state MDP. No such result is known for DEC-POMDPs, making it difficult to generalize value iteration to the multiagent case. This lack of a shared belief-state requires a new set of tools to be developed for solving DEC-POMDP. As a step in this direction, we were able to develop an optimal policy iteration algorithm for DEC-POMDPs that includes the POMDP version as a special case. This algorithm is the focus of the section.

We first show how to extend the definition of a stochastic controller to the multiagent case. Multiagent controllers include a *correlation device*, which is a source of randomness shared by all the agents. This shared randomness increases solution quality while minimally increasing representation size without adding communication. As in the single agent case, policy iteration alternates between exhaustive backups and value-preserving transforma-tions. A convergence proof is given, along with efficient transformations that extend those presented in the previous section.

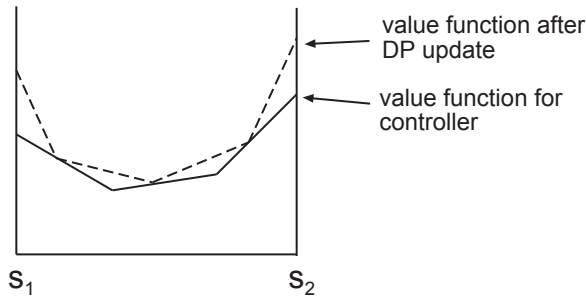

Figure 5: A local optimum for bounded backups. The solid line is the value function for the controller, and the dotted line is the value function for the controller that results from a full DP update.

## 4.1 Correlated Finite-State Controllers

The joint policy for the agents is represented using a stochastic finite-state controller for each agent. In this section, we first define a type of controller in which the agents act independently. We then provide an example demonstrating the utility of correlation, and show how to extend the definition of a controller to allow for correlation among agents.

### 4.1.1 Local Finite-State Controllers

In a local controller, the agent's node is based on the local observations received, and the agent's action is based on the current node. These local controllers are defined in the same way as the POMDP controllers above, with each agent possessing its own controller that operates independently of the others. As before, stochastic transitions and action selection are allowed.

We formally define a *local controller* for agent $i$ as a tuple $\langle Q_i, \Omega_i, A_i, \psi_i, \eta_i \rangle$, where

- $Q_i$ is a finite set of controller nodes.

- $\Omega_i$ is a set of inputs, taken to be the local observations for agent $i$.

- $A_i$ is a set of outputs, taken to be the actions for agent $i$.

- $\psi_i : Q_i \to \Delta A_i$ is an action selection function for agent $i$, defining the distribution of actions selected at each node of that agent's controller.

- $\eta_i : Q_i \times A_i \times \Omega_i \to \Delta Q_i$ is a transition function for agent $i$, defining the distribution of resulting nodes for each initial node and action taken of that agent's controller.

The functions $\psi_i$ and $\eta_i$ parameterize the conditional distribution $P(a_i, q_i'|q_i, o_i)$ which represents the combined action selection and node transition probability for agent $i$. When taken together, the agents' controllers determine the conditional distribution $P(\vec{a}, \vec{q}' | \vec{q}, \vec{o})$.
This is denoted an *independent joint controller*. In the following subsection, we show that
independence can be limiting.

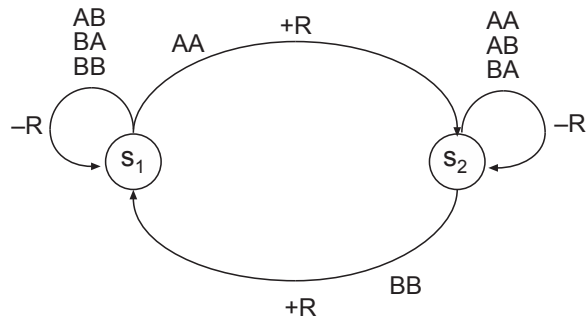

Figure 6: A DEC-POMDP for which a correlated joint policy yields more reward than the
optimal independent joint policy.

### 4.1.2 The Utility of Correlation

The joint controllers described above do not allow the agents to correlate their behavior
via a shared source of randomness. We will use a simple example to illustrate the utility
of correlation in partially observable domains where agents have limited memory. This
example generalizes the one given by Singh (1994) to illustrate the utility of stochastic
policies in partially observable settings containing a single agent.

Consider the DEC-POMDP shown in Figure 6. This problem has two states, two agents,
and two actions per agent ($A$ and $B$). The agents each have only one observation, and
thus cannot distinguish between the two states. For this example, we will consider only
memoryless policies.

Suppose that the agents can independently randomize their behavior using distributions
$P(a_1)$ and $P(a_2)$. If the agents each choose either $A$ or $B$ according to a uniform distribution,
then they receive an expected reward of $-\frac{R}{2}$ per time step, and thus an expected long-term
reward of $\frac{-R}{2(1-\beta)}$. It is straightforward to show that no independent policy yields higher
reward than this one for all states.

Next, let us consider the even larger class of policies in which the agents may act in a
correlated fashion. In other words, we consider all joint distributions $P(a_1, a_2)$. Consider
the policy that assigns probability $\frac{1}{2}$ to the pair $AA$ and probability $\frac{1}{2}$ to the pair $BB$. This
yields an average reward of 0 at each time step and thus an expected long-term reward of
0. The difference between the rewards obtained by the independent and correlated policies
can be made arbitrarily large by increasing $R$.

### 4.1.3 Correlated Joint Controllers

In the previous subsection, we established that correlation can be useful in the face of limited memory. In this subsection, we extend our definition of a joint controller to allow for correlation among the agents. To do this, we introduce an additional finite-state machine, called a correlation device, that provides extra signals to the agents at each time step. The device operates independently of the DEC-POMDP process, and thus does not provide agents with information about the other agents' observations. In fact, the random numbers necessary for its operation could be determined prior to execution time and made available to all agents.

Formally, a *correlation device* is a tuple $\langle Q_c, \psi_c \rangle$, where $Q_c$ is a set of nodes and $\psi_c : Q_c \rightarrow \Delta Q_c$ is a state transition function. At each step, the device undergoes a transition, and each agent observes its state.

We must modify the definition of a local controller to take the state of the correlation device as input. Now, a local controller for agent $i$ is a conditional distribution of the form $P(a_i, q_i'|q_c, q_i, o_i)$. The correlation device together with the local controllers form a joint conditional distribution $P(\vec{a}, \vec{q}'|\vec{q}, \vec{o})$, where $\vec{q} = \langle q_c, q_1, \ldots, q_n \rangle$. We will refer to this as a *correlated joint controller*. Note that a correlated joint controller with $|Q_c| = 1$ is effectively an independent joint controller. Figure 7 contains a graphical representation of the probabilistic dependencies in a correlated joint controller.

The value function for a correlated joint controller can be computed by solving the following system of linear equations, one for each $s \in S$ and $\vec{q} \in \vec{Q}$:

$$V(s, \vec{q}) = \sum_{\vec{a}} P(\vec{a}|\vec{q}) \left[ R(s, \vec{a}) + \beta \sum_{s', \vec{o}, \vec{q}'} P(s', \vec{o}|s, \vec{a}) P(\vec{q}'|\vec{q}, \vec{a}, \vec{o}) V(s', \vec{q}') \right].$$

We sometimes refer to the value of the controller for an initial state distribution. For a distribution $b$, this is defined as

$$V(b) = \max_{\vec{q}} \sum_{s} b(s) V(s, \vec{q}).$$

It is assumed that, given an initial state distribution, the controller is started in the joint node which maximizes value from that distribution.

It is worth noting that correlation can increase the value of a set of fixed-size controllers, but this same value can be achieved by a larger set of uncorrelated controllers. Thus, correlation is a way to make better use of limited representation size, but is not required to produce a set of optimal controllers. This is formalized by the following theorem, which is proved in Appendix A. The theorem asserts the existence of uncorrelated controllers; determining how much extra memory is needed to replace a correlation device remains an open problem.

**Theorem 1** *Given an initial state and a correlated joint controller, there always exists some finite-size joint controller without a correlation device that produces at least the same value for the initial state.*

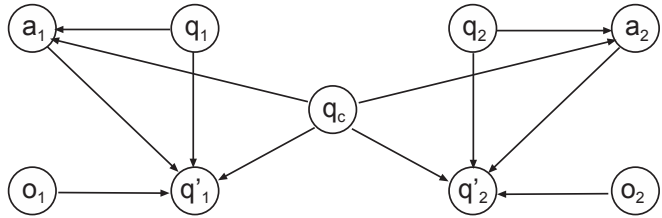

Figure 7: A graphical representation of the probabilistic dependencies in a correlated joint controller for two agents.

In the example above, higher value can be achieved with two node uncorrelated controllers for each agent. If the problem starts in $s_1$, the first node for each agent would choose A and transition to the second node which would choose B. The second node would then transition back to the first node. The resulting policy consists of the agents alternating between choosing AA and BB, producing an expected long-term reward of $\frac{R}{1-\beta}$ which is higher than the correlated one node policy value of 0. Thus, doubling memory for each agent in this problem is sufficient to remove the correlation device.

## 4.2 Policy Iteration

In this section, we describe the policy iteration algorithm. We first extend the definitions of exhaustive backup and value-preserving transformation to the multiagent case. Following that, we provide a description of the complete algorithm, along with a convergence proof.

### 4.2.1 Exhaustive Backups

We introduced exhaustive backups in the section on dynamic programming for POMDPs. We stated that one way to implement a DP update was to perform an exhaustive backup, and then prune dominated nodes that were created. More efficient implementations were described thereafter. These implementations involved interleaving pruning with node generation.

For the multiagent case, it is an open problem whether pruning can be interleaved with node generation. Nodes can be removed, as we will show in a later subsection, but for convergence we require exhaustive backups. We do not define DP updates for the multiagent case, and instead make exhaustive backups a central component of our algorithm.

An exhaustive backup adds nodes to the local controllers for all agents at once, and leaves the correlation device unchanged. For each agent $i$, $|A_i||Q_i|^{|\Omega_i|}$ nodes are added to the local controller, one for each one-step policy. Thus, the joint controller grows by $|Q_c|\prod_i |A_i||Q_i|^{|O_i|}$ joint nodes.

Note that repeated application of exhaustive backups amounts to a brute force search in the space of deterministic policies. This converges to optimality, but is obviously quite inefficient. As in the single agent case, we must modify the joint controller in between adding new nodes. For convergence, these modifications must preserve value in a sense that will be made formal in the following section.

### 4.2.2 VALUE-PRESERVING TRANSFORMATIONS

We now extend the definition of a value-preserving transformation to the multiagent case. In the following subsection, we show how this definition allows for convergence to optimality as the number of iterations grows.

The dual interpretation of dominance is helpful in understanding multiagent value-preserving transformations. Recall that for a POMDP, we say that a node is dominated if there is a convex combination of other nodes with value at least as high for all states. Though we defined a value-preserving transformation in terms of the value function across belief states, we could have equivalently defined it so that every node in the original controller has a dominating convex combination in the new controller.

For the multiagent case, we do not have the concept of a belief state MDP, so we take the second approach mentioned above. In particular, we require that dominating convex combinations exist for nodes of all the local controllers and the correlation device. A transformation of a controller $C$ to a controller $D$ qualifies as a *value-preserving transformation* if $C \leq D$, where $\leq$ is defined below.

Consider correlated joint controllers $C$ and $D$ with node sets $\vec{Q}$ and $\vec{R}$, respectively. We say that $C \leq D$ if there exist mappings $f_i : Q_i \to \Delta R_i$ for each agent $i$ and $f_c : Q_c \to \Delta R_c$ such that

$$V(s, \vec{q}) \leq \sum_{\vec{r}} P(\vec{r}|\vec{q}) V(s, \vec{r})$$

for all $s \in S$ and $\vec{q} \in \vec{Q}$. Note that this relation is transitive as further value-preserving transformations of $D$ will also be value-preserving transformations of $C$.

We sometimes describe the $f_i$ and $f_c$ as a single mapping $f : \vec{Q} \to \Delta \vec{R}$. Examples of efficient value-preserving transformations are given in a later section. In the following subsection, we show that alternating between exhaustive backups and value-preserving transformations yields convergence to optimality.

### 4.2.3 ALGORITHMIC FRAMEWORK

The policy iteration algorithm is initialized with an arbitrary correlated joint controller. In the first part of an iteration, the controller is evaluated via the solution of a system of linear equations. Next, an exhaustive backup is performed to add nodes to the local controllers. Finally, value-preserving transformations are performed.

In contrast to the single agent case, there is no Bellman residual for testing convergence to $\epsilon$-optimality. We resort to a simpler test for $\epsilon$-optimality based on the discount rate and the number of iterations so far. Let $|R_{\max}|$ be the largest absolute value of an immediate reward possible in the DEC-POMDP. Our algorithm terminates after iteration $t$ if $\frac{\beta^{t+1}|R_{\max}|}{1-\beta} \leq \epsilon$. At this point, due to discounting, the value of any policy after step $t$ is less than $\epsilon$. Justification for this test is provided in the convergence proof. The complete algorithm is sketched in Table 5.

Before proving convergence, we state a key lemma regarding the ordering of exhaustive backups and value-preserving transformations. Its proof is deferred to the Appendix.

---

Input: A correlated joint controller, and a parameter $\epsilon$.

1. Evaluate the correlated joint controller by solving a system of linear equations.

2. Perform an exhaustive backup to add deterministic nodes to the local controllers.

3. Perform value-preserving transformations on the controller.

4. If $\frac{\beta^{t+1}|R_{\max}|}{1-\beta} \le \epsilon$, where $t$ is the number of iterations so far, then terminate. Else go to step 1.

Output: A correlated joint controller that is $\epsilon$-optimal for all states.

---

Table 5: Policy Iteration for DEC-POMDPs.

**Lemma 1** *Let $C$ and $D$ be correlated joint controllers, and let $\hat{C}$ and $\hat{D}$ be the results of performing exhaustive backups on $C$ and $D$, respectively. Then $\hat{C} \le \hat{D}$ if $C \le D$.*

Thus, if there is a value-preserving transformation mapping controller $C$ to $D$ and both are exhaustively backed up, then there is a value-preserving transformation mapping controller $\hat{C}$ to $\hat{D}$. This allows value-preserving transformations to be performed before exhaustive backups, while ensuring that value is not lost after the backup. We can now state and prove the main convergence theorem for policy iteration.

**Theorem 2** *For any $\epsilon$, policy iteration returns a correlated joint controller that is $\epsilon$-optimal for all initial states in a finite number of iterations.*

**Proof:** Repeated application of exhaustive backups amounts to a brute force search in the space of deterministic joint policies. Thus, after $t$ exhaustive backups, the resulting controller is optimal for $t$ steps from any initial state. Let $t$ be an integer large enough that $\frac{\beta^{t+1}|R_{\max}|}{1-\beta} \le \epsilon$. Then any possible discounted sum of rewards after $t$ time steps is small enough that optimality over $t$ time steps implies $\epsilon$-optimality over the infinite horizon.

Now recall the above lemma, which states that performing value-preserving transformations before a backup provides at least as much value as just performing a backup. By an inductive argument, performing $t$ steps of policy iteration is a value-preserving transformation of the result of $t$ exhaustive backups. We have argued that for large enough $t$, the value of the controller resulting from $t$ exhaustive backups is within $\epsilon$ of optimal for all states. Thus, the result of $t$ steps of policy iteration is also within $\epsilon$ of optimal for all states. □

## 4.3 Efficient Value-Preserving Transformations

In this section, we describe how to extend controller reductions and bounded backups to the multiagent case. We will show that both of these operations are value-preserving transformations.

### 4.3.1 CONTROLLER REDUCTIONS

Recall that in the single agent case, a node can be removed if for all belief states, there is another node with value at least as high. The equivalent dual interpretation is that a node can be removed is there exists a convex combination of other nodes with value at least as high across the entire state space.

Using the dual interpretation, we can extend this to a rule for removing nodes in the multiagent case. The rule applies to removing nodes either from a local controller or from the correlation device. Intuitively, in considering the removal of a node from a local controller or the correlation device, we consider the nodes of the other controllers to be part of the hidden state.

More precisely, suppose we are considering removing node $q_i$ from agent $i$'s local controller. To do this, we need to find a distribution $P(\hat{q}_i)$ over nodes $\hat{q}_i \in Q_i \setminus q_i$ such that for all $s \in S$, $q_{-i} \in Q_{-i}$, and $q_c \in Q_c$,

$$V(s, q_i, q_{-i}, q_c) \leq \sum_{\hat{q}_i} P(\hat{q}_i) V(s, \hat{q}_i, q_{-i}, q_c).$$

where $Q_{-i}$ represents the set of nodes for the other agents. Finding such a distribution can be formulated as a linear program, as shown in Table 6a. In this case, success is finding parameters such that $\epsilon \geq 0$. The linear program is polynomial in the sizes of the DEC-POMDP and controllers, but exponential in the number of agents.

If we are successful in finding parameters that make $\epsilon \geq 0$, then we can merge the dominated node into the convex combination of other nodes by changing all incoming links to the dominated controller node to be redirected based on the distribution $P(\hat{q}_i)$. At this point, there is no chance of ever transitioning into $q_i$, and thus it can be removed.

The rule for the correlation device is very similar. Suppose that we are considering the removal of node $q_c$. In this case, we need to find a distribution $P(\hat{q}_c)$ over nodes $\hat{q}_c \in Q_c \setminus q_c$ such that for all $s \in S$ and $\vec{q} \in \vec{Q}$,

$$V(s, \vec{q}, q_c) \leq \sum_{\hat{q}_c} P(\hat{q}_c) V(s, \vec{q}, \hat{q}_c).$$

Note that we abuse notation here and use $\vec{Q}$ for the set of tuples of local controller nodes, excluding the nodes for the correlation device. As in the previous case, finding parameters can done using linear programming. This is shown in Table 6b. This linear program is also polynomial in the the sizes of the DEC-POMDP and controllers, but exponential in the number of agents.

We have the following theorem, which states that controller reductions are value-preserving transformations.

**Theorem 3** *Any controller reduction applied to either a local node or a node of the correlation device is a value-preserving transformation.*

**Proof:** Suppose that we have replaced an agent $i$ node $q_i$ with a distribution over nodes in $Q_i \setminus q_i$. Let us take $f_i$ to be the identity map for all nodes except $q_i$, which will map to the new distribution. We take $f_c$ to be the identity map, and we take $f_j$ to be the identity map for all $j \neq i$. This yields a complete mapping $f$. We must now show that $f$ satisfies the condition given in the definition of a value-preserving transformation.

---

(a) Variables: $\epsilon, x(\hat{q}_i)$

Objective: Maximize $\epsilon$

Improvement constraints:

$$\forall s, q_{-i}, q_c \quad V(s, q_i, q_{-i}, q_c) + \epsilon \leq \sum_{\hat{q}_i} x(\hat{q}_i) V(s, \hat{q}_i, q_{-i}, q_c)$$

Probability constraints:

$$\sum_{\hat{q}_i} x(\hat{q}_i) = 1, \quad \forall \hat{q}_i \quad x(\hat{q}_i) \geq 0$$

---

(b) Variables: $\epsilon, x(q_c)$

Objective: Maximize $\epsilon$

Improvement constraints:

$$\forall s, \vec{q} \quad V(s, \vec{q}, q_c) + \epsilon \leq \sum_{\hat{q}_c} x(\hat{q}_c) V(s, \vec{q}, \hat{q}_c)$$

Probability constraints:

$$\sum_{\hat{q}_c} x(\hat{q}_c) = 1, \quad \forall \hat{q}_c \quad x(\hat{q}_c) \geq 0$$

---

Table 6: (a) The linear program to be solved to find a replacement for agent $i$'s node $q_i$. The variable $x(\hat{q}_i)$ represents $P(\hat{q}_i)$. (b) The linear program to be solved to find a replacement for the correlation node $q_c$. The variable $x(\hat{q}_c)$ represents $P(\hat{q}_c)$.

Let $V_o$ be the value function for the original controller, and let $V_n$ be the value function for the controller with $q_i$ removed. A controller reduction requires that

$$V_o(s, \vec{q}) \leq \sum_{\vec{r}} P(\vec{r}|\vec{q}) V_o(s, \vec{r})$$

for all $s \in S$ and $\vec{q} \in \vec{Q}$. Thus, we have

$$
\begin{aligned}
V_o(s, \vec{q}) &= \sum_{\vec{a}} P(\vec{a}|\vec{q}) \left[ R(s, a) + \beta \sum_{s', \vec{o}, \vec{q}'} P(\vec{q}'|\vec{q}, \vec{a}, \vec{o}) P(s', \vec{o}|s, \vec{a}) V_o(s', \vec{q}') \right] \\
&\leq \sum_{\vec{a}} P(\vec{a}|\vec{q}) \left[ R(s, a) + \beta \sum_{s', \vec{o}, \vec{q}'} P(\vec{q}'|\vec{q}, \vec{a}, \vec{o}) P(s', \vec{o}|s, \vec{a}) \sum_{\vec{r}'} P(\vec{r}'|\vec{q}) V_o(s', \vec{r}') \right]
\end{aligned}
$$

$$= \sum_{\vec{a}} P(\vec{a}|\vec{q}) \left[ R(s,a) + \beta \sum_{s',\vec{o},\vec{q}',\vec{r}'} P(\vec{q}'|\vec{q},\vec{a},\vec{o}) P(s',\vec{o}|s,\vec{a}) P(\vec{r}|\vec{q}) V_o(s,\vec{r}') \right]$$

for all $s \in S$ and $\vec{q} \in \vec{Q}$. Notice that the formula on the right is the Bellman operator for the new controller, applied to the old value function. Denoting this operator $T_n$, the system of inequalities implies that $T_n V_o \geq V_o$. By monotonicity, we have that for all $k \geq 0$, $T_n^{k+1}(V_o) \geq T_n^k(V_o)$. Since $V_n = \lim_{k \to \infty} T_n^k(V_o)$, we have that $V_n \geq V_o$. This is sufficient for $f$ to satisfy the condition in the definition of value-preserving transformation.

The argument for removing a node of the correlation device is almost identical to the one given above. □

### 4.3.2 Bounded Dynamic Programming Updates

In the previous section, we described a way to reduce the size of a controller without sacrificing value. Recall that in the single agent case, we could also use bounded backups to increase the value of the controller while keeping its size fixed. This technique can be extended to the multiagent case. As in the previous section, the extension relies on improving a single local controller or the correlation device, while viewing the nodes of the other controllers as part of the hidden state.

We first describe in detail how to improve a local controller. To do this, we choose an agent $i$, along with a node $q_i$. Then, for each $o_i \in \Omega_i$, we search for new parameters for the conditional distribution $P(a_i, q_i'|q_i, o_i)$.

The search for new parameters works as follows. We assume that the original controller will be used from the second step on, and try to replace the parameters for $q_i$ with better ones for just the first step. In other words, we look for parameters satisfying the following inequality:

$$V(s,\vec{q}) \leq \sum_{\vec{a}} P(\vec{a}|\vec{q}) \left[ R(s,a) + \beta \sum_{s',\vec{o},\vec{q}'} P(\vec{q}'|\vec{q},\vec{a},\vec{o}) P(s',\vec{o}|s,\vec{a}) V(s',\vec{q}') \right]$$

for all $s \in S$, $q_{-i} \in Q_{-i}$, and $q_c \in Q_c$. The search for new parameters can be formulated as a linear program, as shown in Table 7a. Its size is polynomial in the sizes of the DEC-POMDP and the joint controller, but exponential in the number of agents.

The procedure for improving the correlation device is very similar to the procedure for improving a local controller. We first choose a device node $q_c$, and consider changing its parameters for just the first step. We look for parameters satisfying the following inequality:

$$V(s,\vec{q}) \leq \sum_{\vec{a}} P(\vec{a}|\vec{q}) \left[ R(s,a) + \beta \sum_{s',\vec{o},\vec{q}'} P(\vec{q}'|\vec{q},\vec{a},\vec{o}) P(s',\vec{o}|s,\vec{a}) V(s',\vec{q}') \right]$$

for all $s \in S$ and $\vec{q} \in \vec{Q}$.

As in the previous case, the search for parameters can be formulated as a linear program. This is shown in Table 7b. This linear program is also polynomial in the sizes of the DEC-POMDP and joint controller, but exponential in the number of agents.

The following theorem states that bounded backups preserve value.

(a) Variables: $\epsilon, x(q_c, a_i), x(q_c, a_i, o_i, q_i')$

Objective: Maximize $\epsilon$

Improvement constraints:

$$\forall s, q_{-i}, q_c \quad V(s, \vec{q}, q_c) + \epsilon \;\; \leq \;\; \sum_{\vec{a}} P(a_{-i}|q_c, q_{-i})[x(q_c, a_i)R(s, \vec{a}) +$$
$$\beta \sum_{s', \vec{o}, \vec{q}', q_c'} x(c, a_i, o_i, q_i')P(q_{-i}'|q_c, q_{-i}, a_{-i}, o_{-i})$$
$$\cdot P(\vec{o}, s'|s, \vec{a})P(q_c'|q_c)V(s', \vec{q}', q_c')]$$

Probability constraints:

$$\forall q_c \quad \sum_{a_i} x(q_c, a_i) = 1, \quad \forall q_c, a_i, o_i \quad \sum_{q_i'} x(q_c, a_i, o_i, q_i') = x(q_c, a_i)$$

$$\forall q_c, a_i \quad x(q_c, a_i) \geq 0, \quad \forall q_c, a_i, o_i, q_i' \quad x(q_c, a_i, o_i, q_i') \geq 0$$

---

(b) Variables: $\epsilon, x(q_c')$

Objective: Maximize $\epsilon$

Improvement constraints:

$$\forall s, \vec{q} \quad V(s, \vec{q}, q_c) + \epsilon \;\; \leq \;\; \sum_{\vec{a}} P(\vec{a}|q_c, \vec{q})[R(s, \vec{a}) + \beta \sum_{s', \vec{o}, \vec{q}', q_c'} P(\vec{q}'|q_c, \vec{q}, \vec{a}, \vec{o})$$
$$\cdot P(s', \vec{o}|s, \vec{a})x(q_c')V(s', \vec{q}', q_c')]$$

Probability constraints:

$$\forall q_c' \quad \sum_{q_c'} x(q_c') = 1, \quad \forall q_c' \quad x(q_c') \geq 0$$

Table 7: (a) The linear program used to find new parameters for agent $i$'s node $q_i$. The variable $x(q_c, a_i)$ represents $P(a_i|q_i, q_c)$, and the variable $x(q_c, a_i, o_i, q_i')$ represents $P(a_i, q_i'|q_c, q_i, o_i)$. (b) The linear program used to find new parameters for the correlation device node $q_c$. The variable $x(q_c')$ represents $P(q_c'|q_c)$.

**Theorem 4** *Performing a bounded backup on a local controller or the correlation device produces a new correlated joint controller which is a value-preserving transformation of the original.*

**Proof:** Consider the case in which some node $q_i$ of agent $i$'s local controller is changed. We define $f$ to be a deterministic mapping from nodes in the original controller to the corresponding nodes in the new controller.

Let $V_o$ be the value function for the original controller, and let $V_n$ be the value function for the new controller. Recall that the new parameters for $P(a_i, q_i'|q_c, q_i, o_i)$ must satisfy the following inequality for all $s \in S$, $q_{-i} \in Q_{-i}$, and $q_c \in Q_c$:

$$V_o(s, \vec{q}) \leq \sum_{\vec{a}} P(\vec{a}|\vec{q}) \left[ R(s, a) + \beta \sum_{s', \vec{o}, \vec{q}'} P(\vec{q}'|\vec{q}, \vec{a}, \vec{o}) P(s', \vec{o}|s, \vec{a}) V_o(s', \vec{q}') \right].$$

Notice that the formula on the right is the Bellman operator for the new controller, applied to the old value function. Denoting this operator $T_n$, the system of inequalities implies that $T_n V_o \geq V_o$. By monotonicity, we have that for all $k \geq 0$, $T_n^{k+1}(V_o) \geq T_n^k(V_o)$. Since $V_n = \lim_{k \to \infty} T_n^k(V_o)$, we have that $V_n \geq V_o$. Thus, the new controller is a value-preserving transformation of the original one.

The argument for changing nodes of the correlation device is almost identical to the one given above. □

## 4.4 Open Issues

We noted at the beginning of the section that there is no known way to convert a DEC-POMDP into an equivalent belief-state MDP. Despite this fact, we were able to develop a provably convergent policy iteration algorithm. However, the policy iteration algorithm for POMDPs has other desirable properties besides convergence, and we have not yet been able to extend these to the multiagent case. Two such properties are described below.

### 4.4.1 Error Bounds

The first property is the existence of a Bellman residual. In the single agent case, it is possible to compute a bound on the distance to optimality using two successive value functions. In the multiagent case, policy iteration produces a sequence of controllers, each of which has a value function. However, we do not have a way to obtain an error bound from these value functions. For now, to bound the distance to optimality, we must consider the discount rate and the number of iterations completed.

### 4.4.2 Avoiding Exhaustive Backups

In performing a DP update for POMDPs, it is possible to remove certain nodes from consideration without first generating them. In Section 3, we gave a high-level description of a few different approaches to doing this. For DEC-POMDPs, however, we did not define a DP update and instead used exhaustive backups as the way to expand a controller. Since exhaustive backups are expensive, it would be useful to extend the more sophisticated pruning methods for POMDPs to the multiagent case.

---

Input: A joint controller, the desired number of centralized belief points $k$, initial state $b_0$ and fixed policy for each agent $\pi_i$.

1. Starting from $b_0$, sample a set of $k$ belief points for each agent assuming the other agents use their fixed policy.

2. Evaluate the joint controller by solving a system of linear equations.

3. Perform an exhaustive backup to add deterministic nodes to the local controllers.

4. Retain nodes that contribute the highest value at each of the belief points.

5. For each agent, replace nodes that have lower value than some combination of other nodes at each belief point.

6. If controller sizes and parameters do not change then terminate. Else go to step 2.

Output: A new joint controller based on the sampled centralized belief points.

---

Table 8: Heuristic Policy Iteration for DEC-POMDPs.

Unfortunately, in the case of POMDPs, the proofs of correctness for these methods all use the fact that there exists a Bellman equation. Roughly speaking, this equation allows us to determine whether a potential node is dominated by just analyzing the nodes that would be its successors. Because we do not currently have an analog of the Bellman equation for DEC-POMDPs, we have not been able to generalize these results.

There is one exception to the above statement, however. When an exhaustive backup has been performed for all agents except one, then a type of belief state space can be constructed for the agent in question using the system states and the nodes for the other agents. The POMDP node generation methods can then be applied to just that agent. In general, though, it seems difficult to rule out a node for one agent before generating all the nodes for the other agents.

## 5. Heuristic Policy Iteration

While the optimal policy iteration method shows how a set of controllers with value arbitrarily close to optimal can be found, the resulting controllers may be very large and many unnecessary nodes may be generated along the way. This is exacerbated by the fact that the algorithm cannot take advantage of an initial state distribution and must attempt to improve the controller for any initial state. As a way to combat these disadvantages, we have developed a heuristic version of policy iteration that removes nodes based on their value only at a given set of centralized belief points. We call these centralized belief points because they are distributions over the system state that in general could only be known by full observability of the problem. As a result, the algorithm will no longer be optimal, but it can often produce more concise controllers with higher solution quality for a given initial state distribution.

## 5.1 Directed Pruning

Our heuristic policy iteration algorithm uses sets of belief points to direct the pruning process of our algorithm. There are two main advantages of this approach: it allows simultaneous pruning for all agents and it focuses the controller on certain areas of the belief space. We first discuss the benefits of simultaneous pruning and then mention the advantages of focusing on small areas of the belief space.

As mentioned above, the pruning method used by the optimal algorithm will not always remove all nodes that could be removed from all the agents' controllers without losing value. Because pruning requires each agent to consider the controllers of other agents, after nodes are removed for one agent, the other agents may be able to prune other nodes. Thus pruning must cycle through the agents and ceases when no agent can remove any further nodes. This is both time consuming and causes the controller to be much larger than it needs to be.

Like the game theoretic concept of incredible threats[1], a set of suboptimal policies for an agent may be useful only because other agents may employ similarly suboptimal policies. That is, because pruning is conducted for each agent while holding the other agents' policies fixed, polices that are useful for any set of other agent policies are retained, no matter the quality of these other agent policies. Some of an agent's policies may only be retained because they have the highest value when used in conjunction with other suboptimal policies of the other agents. In these cases, only by removing the set of suboptimal policies simultaneously can controller size be reduced while at least maintaining value. This simultaneous pruning could further reduce controller sizes and thus increase scalability and solution quality. While it may be possible to define a value-preserving transformation for these problems, finding a nontrivial automated way to do so while maintaining the optimality of the algorithm remains an open question.

The advantage of considering a smaller part of the state space has already been shown to produce drastic performance increases in POMDPs (Ji, Parr, Li, Liao, & Carin, 2007; Pineau, Gordon, & Thrun, 2003) and finite-horizon DEC-POMDPs (Seuken & Zilberstein, 2007; Szer & Charpillet, 2006). For POMDPs, a problem with many states has a belief space with large dimensionality, but many parts may never be visited by an optimal policy. Focusing on a subset of belief states can allow a large part of the state space to be ignored without significant loss of solution quality.

The problem of having a large state space is compounded in the DEC-POMDP case. Not only is there uncertainty about the state, but also about the policies of the other agents. As a consequence, the generalized belief space which includes all possible distributions over states of the system and current policies of the other agents must be considered to guarantee optimality. This results in a huge space which contains many unlikely states and policies. The uncertainty about which policies other agents may utilize does not allow belief updates to normally be calculated for DEC-POMDPs, but as we showed above, it can be done by assuming a probability distribution over actions of the other agents. This limits the number of policies that need to be considered by all agents and if the distributions are chosen well, may permit a high-valued solution to be found.

---

1. An incredible threat is an irrational strategy that the agent knows it will receive a lower value by choosing it. While it is possible the agent will choose the incredible threat strategy, it is irrational to do so.

---

Variables: $\epsilon$, $x(\hat{q}_i)$ and for each belief point $b$

Objective: Maximize $\epsilon$

Improvement constraints: $\quad \forall b, q_{-i} \; \sum_s b(s) \left[ \sum_{\hat{q}_i} x(\hat{q}_i) V(\hat{q}_i, q_{-i}, s) - V(\vec{q}, s) \right] \geq \epsilon$

Probability constraints: $\quad \sum_{\hat{q}_i} x(\hat{q}_i) = 1$ and $\forall \hat{q}_i \; x(q_i) \geq 0$

---

Table 9: The linear program used to determine if a node $q$ for agent $i$ is dominated at each point $b$ and all initial nodes of the other agents' controllers. As node $q$ may be dominated by a distribution of nodes, variable $x(\hat{q}_i)$ represents $P(\hat{q}_i)$, the probability of starting in node $\hat{q}$ for agent $i$.

## 5.2 Belief Set Generation

As mentioned above, our heuristic policy iteration algorithm constructs sets of belief points for each agent which are later used to evaluate the joint controller and remove dominated nodes. To generate the belief point set, we start at the initial state and by making assumptions about the other agents, we can calculate the resulting belief state for each action and observation pair of an agent. By fixing the policies for the other agents, this belief state update can be calculated in a way very similar to that described for POMDPs in section 3.1.1. This procedure can be repeated from each resulting belief state until a desired number of points is generated or no new points are visited.

More formally, we assume the other agents have a fixed distribution of action choice for each system state. That is, if we know $P(\vec{a}_{-i}|s)$ then we can determine the probability any state results given a belief point and an agent's action and observation. The derivation of the likelihood of state $s'$, given the belief state $b$, and agent $i$'s action $a_i$ and observation $o_i$ is shown below.

$$
\begin{aligned}
P(s'|a_i, o_i, b) \;&=\; \sum_{\vec{a}_{-i}, \vec{o}_{-i}, s} P(s', \vec{a}_{-i}, \vec{o}_{-i}, s | a_i, o_i, b) \\
&=\; \frac{\sum_{\vec{a}_{-i}, \vec{o}_{-i}, s} P(\vec{o}|s, b, \vec{a}, s') P(s', s, \vec{a}, b)}{P(o_i, a_i, b)} \\
&=\; \frac{\sum_{\vec{a}_{-i}, \vec{o}_{-i}, s} P(\vec{o}|s, \vec{a}, s') P(s'|s, \vec{a}, b) P(\vec{a}, s, b)}{P(o_i, a_i, b)} \\
&=\; \frac{\sum_{\vec{a}_{-i}, \vec{o}_{-i}, s} P(\vec{o}|s, \vec{a}, s') P(s'|s, \vec{a}) P(\vec{a}_{-i}|a, s, b) P(\vec{a}, s, b)}{P(o_i, a_i, b)} \\
&=\; \frac{\sum_{\vec{a}_{-i}, \vec{o}_{-i}, s} P(\vec{o}|s, \vec{a}, s') P(s'|s, \vec{a}) P(\vec{a}_{-i}|a_i, s, b) P(s|a_i, b) P(a_i, b)}{P(o_i, a_i, b)}
\end{aligned}
$$

$$= \frac{\sum_{\vec{a}_{-i}, \vec{o}_{-i}, s} P(\vec{o}|s, \vec{a}, s') P(s'|s, \vec{a}) P(\vec{a}_{-i}|s) b(s)}{P(o_i|a_i, b)}$$

where

$$P(o_i|a_i, b) = \sum_{a_{-i}, o_{-i}, s, s'} P(\vec{o}|s, \vec{a}, s') P(s'|s, \vec{a}) P(\vec{a}_{-i}|s) b(s)$$

Thus, given the action probabilities for the other agents, $-i$, and the transition and observation models of the system, a belief state update can be calculated.

### 5.3 Algorithmic Framework

We provide a formal description of our approach in Table 8. Given the desired number of belief points, $k$, and random action and observation selection for each agent, the sets of points are generated as described above. The search begins at the initial state of the problem and continues until the given number of points is obtained. If no new points are found, this process can be repeated to ensure a diverse set is produced. The arbitrary initial controller is evaluated and the value at each state and for each initial node of any agent's controller is retained. The exhaustive backup procedure is exactly the same as the one used in the optimal algorithm, but updating the controller takes place in two steps. First, for each of the $k$ belief points, the highest-valued set of initial nodes is found. To accomplish this, the value of beginning at each combination of nodes for all agents is calculated for each of these $k$ points and the best combination is kept. This allows nodes that do not contribute to any of these values to be simultaneously pruned. Next, each node of each agent is pruned using the linear program shown in Table 9. If a distribution of nodes for the given agent has higher value at each of the belief points for any initial nodes of the other agents' controllers, it is pruned and replaced with that distribution. The new controllers are then evaluated and the value is compared with the value of the previous controller. This process of backing up and pruning continues while the controller parameters continue to change.

Similar to how bounded policy updates can be used in conjunction with pruning in the optimal policy iteration algorithm, a nonlinear programming approach (Amato et al., 2007) can be used to improve solution quality for the heuristic case. To accomplish this, instead of optimizing the controller for just the initial belief state of the problem, all the belief points being considered are used. A simple way to achieve this is to maximize over the sum of the values of the initial nodes of the controllers weighted by the probabilities given for each point. This approach can be used after each pruning step and may further improve value of the controllers.

## 6. Dynamic Programming Experiments

This section describes the results of experiments performed using policy iteration. Because of the flexibility of the algorithm, it is impossible to explore all possible ways of implementing it. However, we did experiment with a few different implementation strategies to gain an idea of how the algorithm works in practice. All of these experiments were run on a 3.40GHz Intel Pentium 4 with 2GB of memory. Three main sets of experiments were performed on a single set of test problems.

Our first set of experiments focused on exhaustive backups and controller reductions. The results confirm that value improvement can be obtained through iterated application of these two operations. Further improvement is demonstrated by also incorporating bounded updates. However, because exhaustive backups are expensive, the algorithm was unable to complete more than a few iterations on any of our test problems.

In the second set of experiments, we addressed the complexity issues by using only bounded backups, and no exhaustive backups. With bounded backups, we were able to obtain higher-valued controllers while keeping memory requirements fixed. We examined how the sizes of the initial local controllers and the correlation device affected the value of the final solution.

The third set of experiments examined the complexity issues caused by exhaustive backups by using the point-based heuristic. This allowed our heuristic policy iteration algorithm to complete more iterations than the optimal algorithm and in doing so, increased solution quality of the largest solvable controllers. By incorporating Amato et al.'s NLP approach, the heuristic algorithm becomes slightly less scalable than with heuristic pruning alone, but the amount of value improvement per step increases. This causes the resulting controllers in each domain to have the highest value of any approach.

## 6.1 Test Domains

In this section, we describe three test domains, ordered by the size of the problem representation. For each problem, the transition function, observation function, and reward functions are described. In addition, an initial state is specified. Although policy iteration does not require an initial state as input, one is commonly assumed and is used by the heuristic version of the algorithm. A few different initial states were tried for each problem, and qualitatively similar results were obtained. In all domains, a discount factor of 0.9 was utilized.

As a very loose upper bound, the centralized policy was calculated for each problem in which all agents share their observations with a central agent and decisions for all agents are made by the central agent. This results in a POMDP with the same number of states, but the action and observation sets are Cartesian products of the agents action and observation sets. The value of this POMDP policy is provided below, but because DEC-POMDP policies are more constrained, the optimal value may be much lower.

### Two Agent Tiger Problem

The two agent tiger problem consists of 2 states, 3 actions and 2 observations (Nair et al., 2003). The domain includes two doors, one of which leads to a tiger and the other to a large treasure. Each agent may open one of the doors or listen. If either agent opens the door with the tiger behind it, a large penalty is given. If the door with the treasure behind it is opened and the tiger door is not, a reward is given. If both agents choose the same action (i.e., both opening the same door) a larger positive reward or a smaller penalty is given to reward cooperation. If an agent listens, a small penalty is given and an observation is seen that is a noisy indication of which door the tiger is behind. While listening does not change the location of the tiger, opening a door causes the tiger to be placed behind one of the door with equal probability. The problem begins with the tiger equally likely to be located behind either door. The optimal centralized policy for this problem has value 59.817.

### Meeting on a Grid

In this problem, with 16 states, 5 actions and 4 observations, two robots must navigate on a two-by-two grid. Each robot can only sense whether there are walls to its left or right, and their goal is to spend as much time as possible on the same square as the other agent. The actions are to move up, down, left, or right, or to stay on the same square. When a robot attempts to move to an open square, it only goes in the intended direction with probability 0.6, otherwise it either goes in another direction or stays in the same square. Any move into a wall results in staying in the same square. The robots do not interfere with each other and cannot sense each other. The reward is 1 when the agents share a square, and 0 otherwise. The initial state places the robots diagonally across from each other and the optimal centralized policy for this problem has value 7.129.

### Box Pushing Problem

This problem, with 100 states, 4 actions and 5 observations consists of two agents that get rewarded by pushing different boxes (Seuken & Zilberstein, 2007). The agents begin facing each other in the bottom corners of a four-by-three grid with the available actions of turning right, turning left, moving forward or staying in place. There is a 0.9 probability that the agent will succeed in moving and otherwise will stay in place, but the two agents can never occupy the same square. The middle row of the grid contains one large box in the middle of two small boxes. The small boxes can be moved by a single agent, but the large box can only be moved by both agents pushing at the same time. The upper row of the grid is considered the goal row, which the boxes are pushed into. The possible deterministic observations for each agent consist of seeing an empty space, a wall, the other agent, a small box or the large box. A reward of 100 is given if both agents push the large box to the goal row and 10 is given for each small box that is moved to the goal row. A penalty of -5 is given for each agent that cannot move and -0.1 is given for each time step. Once a box is moved to the goal row, the environment resets to the original start state. The optimal centralized policy for this problem has value 183.936.

## 6.2 Exhaustive Backups and Controller Reductions

In this section, we present the results of using exhaustive backups together with controller reductions. For each domain, the initial controllers for each agent contained a single node with a self loop, and there was no correlation device. For each problem, the first action of the problem description was used. This resulted in the repeated actions of opening the left door in the two agent tiger problem, moving up in the meeting on a grid problem and turning left in the box pushing problem. The reason for starting with the smallest possible controllers was to see how many iterations we could complete before running out of memory.

On each iteration, we performed an exhaustive backup, and then alternated between agents, performing controller reductions until no more nodes could be removed. For bounded dynamic programming results, after the reductions were completed bounded updates were also performed for all agents. For these experiments, we attempted to improve the nodes of each agent in turn until value could not be improved for any node of any agent. For each iteration, we recorded the sizes of the controllers produced, and noted what the sizes would be if no controller reductions had been performed. In addition, we recorded the value from the initial state and the total time taken to reach the given result.

| Two Agent Tiger, $|S| = 2$, $|A_i| = 3$, $|\Omega_i| = 2$ | | | |
|:---:|:---:|:---:|:---:|
| Iteration | Exhaustive Sizes | Controller Reductions | Bounded Updates |
| 0 | $(1, 1)$ | -150 (1,1 in 1s) | -150 (1,1 in 1s) |
| 1 | $(3, 3)$ | -137 (3,3 in 1s) | -20 (3,3 in 12s) |
| 2 | $(27, 27)$ | -117.8 (15, 15 in 7s) | -20 (15, 15 in 89s) |
| 3 | $(2187, 2187)$ | -98.9 (255, 255 in 1301s) | -20* (255, 255 in 3145s) |

| Meeting on a Grid, $|S| = 16$, $|A_i| = 5$, $|\Omega_i| = 4$ | | | |
|:---:|:---:|:---:|:---:|
| Iteration | Exhaustive Sizes | Controller Reductions | Bounded Updates |
| 0 | $(1, 1)$ | 2.8 (1,1 in 1s) | 2.8 (1,1 in 1s) |
| 1 | $(5, 5)$ | 3.4 (5,5 in 7s) | 3.8 (5,5 in 145s) |
| 2 | $(3125, 3125)$ | 3.7 (80,80 in 821s) | 4.78* (125,125 in 1204s) |

| Box Pushing, $|S| = 100$, $|A_i| = 4$, $|\Omega_i| = 5$ | | | |
|:---:|:---:|:---:|:---:|
| Iteration | Exhaustive Sizes | Controller Reductions | Bounded Updates |
| 0 | $(1, 1)$ | -2 (1,1 in 4s) | -2 (1,1 in 53s) |
| 1 | $(4, 4)$ | -2 (2,2 in 108s) | 6.3 (2,2 in 132s) |
| 2 | $(4096, 4096)$ | 12.8 (9,9 in 755s) | 42.7* (16,17 in 714s) |

Table 10: Results of applying exhaustive backups, controller reductions and bounded updates to our test problems. The second column contains the sizes of the controllers if only exhaustive backups had been performed. The third column contains the resulting value, sizes of the controllers, and time required for controller reductions to be performed on each iteration. The fourth column displays these same quantities with bounded updates also being used. The * denotes that a backup and pruning were performed, but bounded updates exhausted the given resources.

The results are shown in Table 10. Because exhaustive backups add many nodes, we were unable to complete many iterations without exceeding memory limits. As expected, the smallest problem led to the largest number of iterations being completed. Although we could not complete many iterations before running out of memory, the use of controller reductions led to significantly smaller controllers compared to the approach of just applying exhaustive backups. Incorporating bounded updates requires some extra time, but is able to improve the value produced at each step, causing substantial improvement in some cases.

It is also interesting to notice that the controller sizes when using bounded updates are not always the same as when only controller reductions are completed. This can be seen after two iterations in both the meeting on a grid and box pushing problems. This can occur because the bounded updates change node value and thus change the number and location of the nodes that are pruned. In the box pushing problem, the two agents also have different size controllers after two steps. This can occur, even in symmetric problems, when a set of actions is only necessary for a single agent.

## 6.3 Bounded Dynamic Programming Updates

As we saw from the previous experiments, exhaustive backups can fill up memory very quickly. This leads naturally to the question of how much improvement is possible without exhaustive backups. In this section, we describe an experiment in which we repeatedly applied bounded backups, which left the size of the controller fixed. We experimented with different starting sizes for the local controllers and the correlation device.

We define a *trial run* of the algorithm as follows. At the start of a trial run, a size is chosen for each of the local controllers and the correlation device. The action selection and transition functions are initialized to be deterministic, with the outcomes drawn according to a uniform distribution. A *step* consists of choosing a node uniformly at random from the correlation device or one of the local controllers, and performing a bounded backup on that node. After 200 steps, the run is considered over. In practice, we found that values often stabilized in fewer steps.

We varied the sizes of the local controllers while maintaining the same number of nodes for each agent, and we varied the size of the correlation device from 1 to 2. For each domain, we increased number of nodes until the required number of steps could not be completed in under four hours. In general, runs required significantly less time to terminate. For each combination of sizes, we performed 20 trial runs and recorded the best value over all runs.

For each of the three problems, we were able to obtain solutions with higher value than with exhaustive backups. Thus, we see that even though repeated application of bounded backups does not have an optimality guarantee, it can be competitive with an algorithm that does. However, it should be noted that we have not performed an exhaustive comparison. We could have made different design decisions for both approaches concerning the starting controllers, the order in which nodes are considered, and other factors.

Besides comparing to the exhaustive backup approach, we wanted to examine the effect of the sizes of the local controllers and the correlation device on value. Figure 8 shows a graph of best values plotted against controller size. We found that, for the most part, the value increases when we increase the size of the correlation device from one node to two nodes (essentially moving from independent to correlated). It is worth noting that the solution quality had somewhat high variance in each problem, showing that setting good initial parameters is important for high-valued solutions.

For small controllers, the best value tends to increase with controller size. However, for very large controllers, this not always the case. This can be explained by considering how a bounded backup works. For new node parameters to be acceptable, they must not decrease the value for any combination of states, nodes for the other controllers, and nodes for the correlation device. This becomes more difficult as the numbers of nodes increase, and thus it is easier to get stuck in a local optimum. This can be readily seen in the two agent tiger problem and to some extent the meeting on a grid problem. Memory was exhausted before this phenomenon takes place in the box pushing problem.

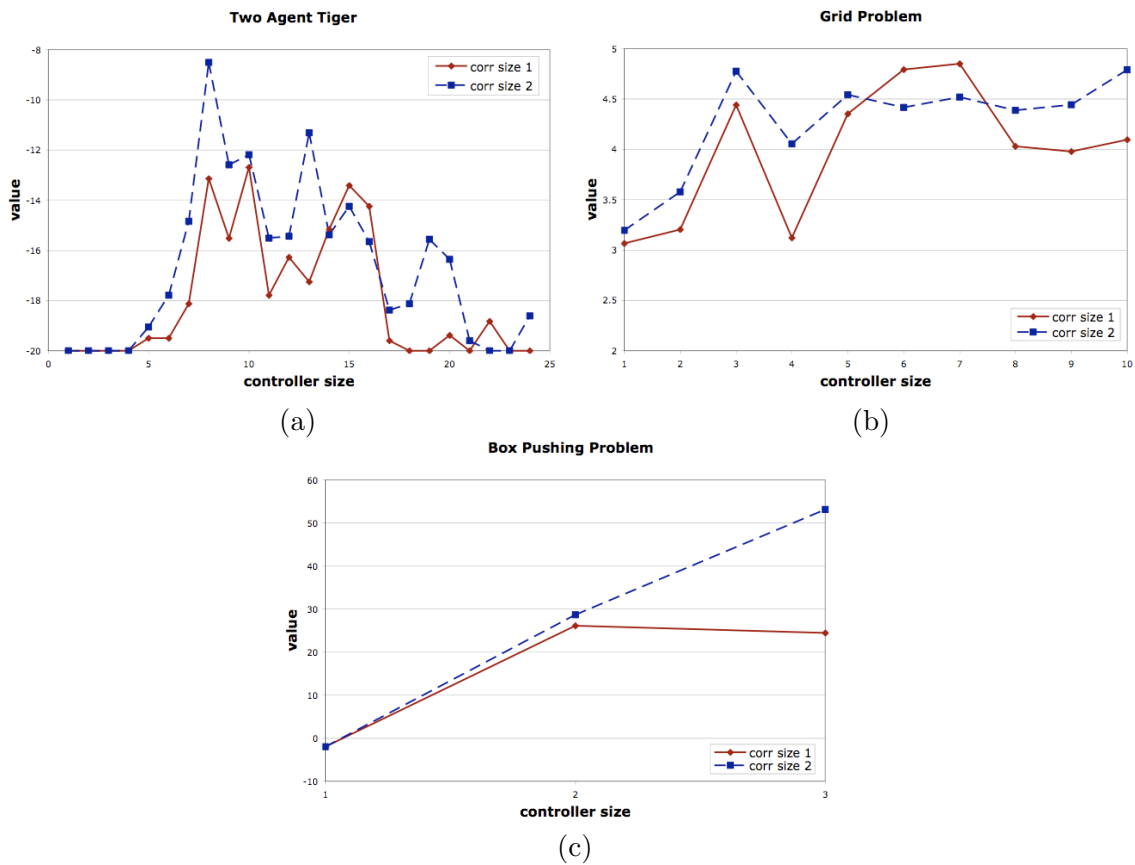

Figure 8: Best value per trial run plotted against the size of the local controllers, for (a) the two agent tiger problem, (b) the meeting in a grid problem and (c) the box pushing problem. The solid line represents independent controllers (a correlation device with one node), and the dotted line represents a joint controller including a two-node correlation device. Times ranged from under 1s for one node controllers without correlation to four hours for the largest controller found with correlation in each problem.

## 6.4 Heuristic Dynamic Programming Updates

As observed above, the optimal dynamic programming approach can only complete a small number of backups before resources are exhausted. Similarly, using bounded updates with fixed size controllers can generate high value solutions, but it can be difficult to pick the correct controller size and initial parameters. As an alternative to the other approaches, we also present experiments using our heuristic dynamic programming algorithm.

Like the optimal policy iteration experiments, we initialized single node controllers for each agent with self loops and no correlation device. The same first actions were used as above and backups were performed until memory was exhausted. The set of belief points for each problem was generated given the initial state distribution and a distribution of actions for the other agents. For the meeting on a grid and box pushing problems, it was assumed that all agents chose any action with equal probability regardless of state. For the two agent tiger problem, it was assumed that for any state agents listen with probability 0.8 and open each door with probability 0.1. This simple heuristic policy was chosen to allow more of the state space to be sampled by our search. The number of belief points used for the two agent tiger and meeting on a grid problems was ten and twenty points were used for the box pushing problem.

For each iteration, we performed an exhaustive backup and then pruned controllers as described in steps four and five of Table 8. All the nodes that contributed to the highest value for each belief point were retained and then each node was examined using the linear program in Table 9. For results with the NLP approach, we also improved the set of controllers after heuristic pruning by optimizing a nonlinear program whose objective was the sum of the values of the initial nodes weighted by the belief point probabilities. We report the value produced by the optimal and heuristic approaches for each iteration that could be completed in under four hours and with the memory limits of the machine used. The nonlinear optimization was performed on the NEOS server, which provides a set of machines with varying CPU speeds and memory limitations.

The values for each iteration of each problem are given in Figure 9. We see the heuristic policy iteration (HPI) methods are able to complete more iterations than the optimal methods and as a consequence produce higher values. In fact, the results from HPI are almost always exactly the same as those for the optimal policy iteration algorithm without bounded updates for all iterations that can be completed by the optimal approach. Thus, improvement occurs primarily due to the larger number of backups that can be performed.

We also see that while incorporating bounded updates improves value for the optimal algorithm, incorporating the NLP approach into the heuristic approach produces even higher value. Optimizing the NLP requires a small time overhead, but substantially increases value on each iteration. This results in the highest controller value in each problem. Using the NLP also allows our heuristic policy iteration to converge to a six node controller for each agent in the two agent tiger problem. Unfortunately, this solution is known to be suboptimal. As an heuristic algorithm, this is not unexpected, and it should be noted that even suboptimal solutions by the heuristic approach outperform all other methods in all our test problems.

## 6.5 Discussion

We have demonstrated how policy iteration can be used to improve both correlated and independent joint controllers. We showed that using controller reductions together with exhaustive backups is more efficient in terms of memory than using exhaustive backups alone. However, due to the complexity of exhaustive backups, even that approach could only complete a few iterations on each of our test problems.

Using bounded backups alone provided a good way to deal with the complexity issues. With bounded backups, we were able to find higher-valued policies than with the previous approach. Through our experiments, we were able to understand how the sizes of the local controllers and correlation device affect the final values obtained.

With our heuristic policy iteration algorithm, we demonstrated further improvement by dealing with some of the complexity issues. The heuristic approach is often able to continue improving solution quality past the point where the optimal algorithm exhausts resources. More efficient use of this limited representation size is achieved by incorporating the NLP approach as well. In fact, the heuristic algorithm with NLP improvements at each step provided results that are at least equal to the highest value obtained in each problem and sometimes were markedly higher than the other approaches. Furthermore, as far as we know, these results are the highest published values for all three of the test domains.

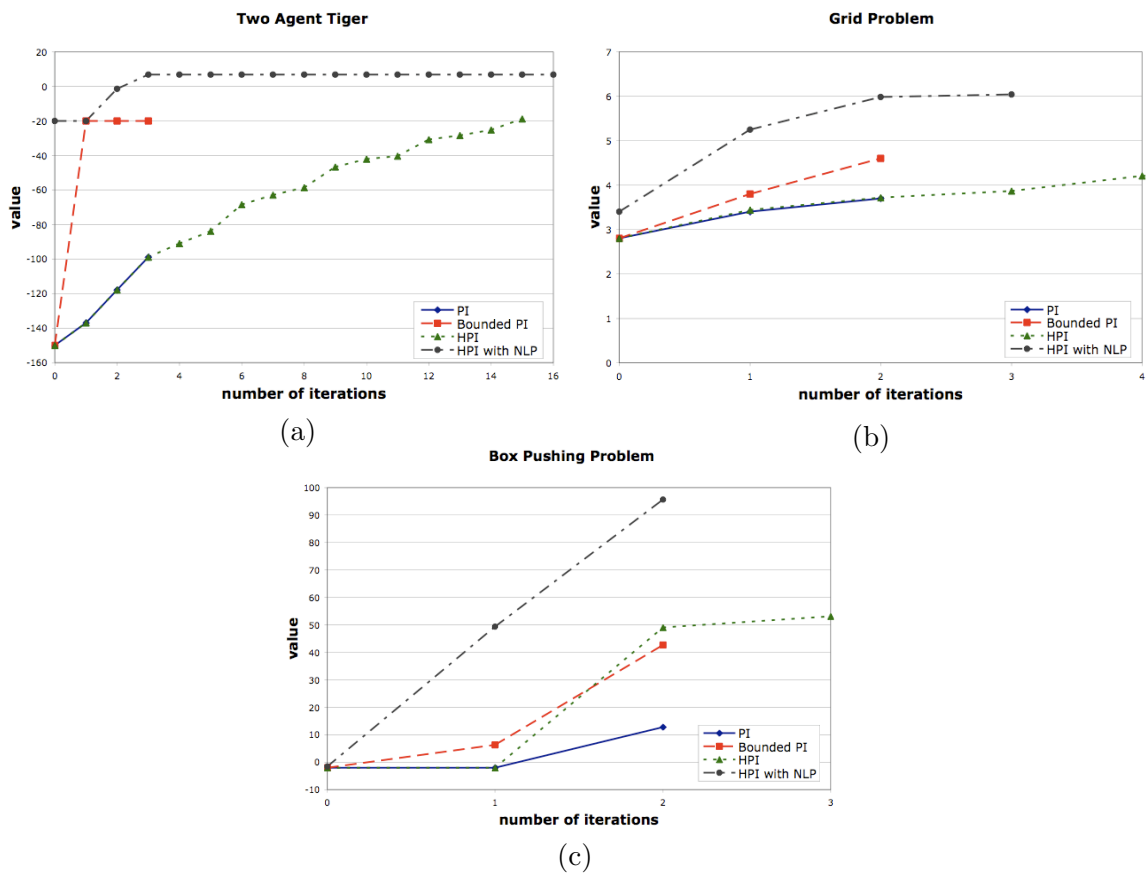

(a)

(b)

(c)

Figure 9: Comparison of the dynamic programming algorithms on (a) the two agent tiger problem, (b) the meeting in a grid problem and (c) the box pushing problem. The value produced by policy iteration with and without bounded backups as well as our heuristic policy iteration with and without optimizing the NLP were compared on each iteration until the time or memory limit was reached.

## 7. Conclusion

We present a policy iteration algorithm for DEC-POMDPs. The algorithm uses a novel policy representation consisting of stochastic finite-state controllers for each agent along with a correlation device. We define value-preserving transformations and show that alternating between exhaustive backups and value-preserving transformations leads to convergence to optimality. We also extend controller reductions and bounded backups from the single agent case to the multiagent case. Both of these operations are value-preserving transformations and are provably efficient. Finally, we introduced a heuristic version of our algorithm which is more scalable and produces higher values on our test problems. Our algorithm serves as the first nontrivial exact algorithm for DEC-POMDPs, and provides a bridge to the large body of work on dynamic programming for POMDPs.

Our work provides a solid foundation for solving DEC-POMDPs, but much work remains in addressing more challenging problem instances. We focused on solving general DEC-POMDPs, but the efficiency of our approaches could be improved by using structure found in certain problems. This would allow specialized representations and solution techniques to be incorporated. Below we describe some key challenges of our general approach, along with some preliminary algorithmic ideas to extend our work on policy iteration.

**Approximation with Error Bounds**   Often, strict optimality requirements cause computational difficulties. A good compromise is to search for policies that are within some bound of optimal. Our framework is easily generalized to allow for this.

Instead of a value-preserving transformation, we could define an $\epsilon$-value-preserving transformation, which insures that the value at all states decreases by at most $\epsilon$. We can perform such transformations with no modifications to any of our linear programs. We simply need to relax the requirement on the value for $\epsilon$ that is returned. It is easily shown that using an $\epsilon$-value-preserving transformation at each step leads to convergence to a policy that is within $\frac{\epsilon \beta}{1-\beta}$ of optimal for all states.

For controller reductions, relaxing the tolerance may lead to smaller controllers because some value can be sacrificed. For bounded backups, it may help in escaping from local optima. Though relaxing the tolerance for a bounded backup could lead to a decrease in value for some states, a small "downward" step could lead to higher value overall in the long run. We are currently working on testing these hypotheses empirically.

**General-Sum Games**   In a general-sum game, there is a set of agents, each with its own set of strategies, and a strategy profile is defined to be a tuple of strategies for all agents. Each agent assigns a payoff to each strategy profile. The agents may be noncooperative, so the same strategy profile may be assigned different values for each agent.

The DEC-POMDP model can be extended to a general-sum game by allowing each agent to have its own reward function. In this case, the strategies are the local policies, and a strategy profile is a joint policy. This model is often called a *partially observable stochastic game (POSG)*. Hansen et al. (2004) presented a dynamic programming algorithm for finite-horizon POSGs. The algorithm was shown to perform iterated elimination of dominated strategies in the game. Roughly speaking, it eliminates strategies that are not useful for an agent, regardless of the strategies of the other agents.

Work remains to be done on extending the notion of a value-preserving transformation to the noncooperative case. One possibility is to redefine value-preserving transformations so that value is preserved for *all* agents. This is closely related to the idea of *Pareto optimality*. In a general-sum game, a strategy profile is said to be Pareto optimal if there does not exist another strategy profile that yields higher payoff for all agents. It seems that policy iteration using the revised definition of value-preserving transformation would tend to move the controller in the direction of the Pareto optimal set. Another possibility is to define value-preserving transformations with respect to specific agents. As each agent transforms its own controller, the joint controller should move towards a Nash equilibrium.

**Handling Large Numbers of Agents** The general DEC-POMDP representation presented in this paper grows exponentially with the number of agents, as seen in the growth of the set of joint actions and observations as well as the transition, reward and observation functions. Thus this representation is not feasible for large numbers of agents. However, a compact representation is possible if each agent interacts directly with just a few other agents. We can have a separate state space for each agent, factored transition probabilities, and a reward function that is the sum of local reward functions for clusters of agents. In this case, the problem size is exponential only in the maximum number of agents interacting directly. This idea is closely related to recent work on graphical games (La Mura, 2000; Koller & Milch, 2003).

Once we have a compact representation, the next question to answer is whether we can adapt policy iteration to work efficiently with the representation. This indeed seems possible. With the value-preserving transformations we presented, the nodes of the other agents are considered part of the hidden state of the agent under consideration. These techniques modify the controller of the agent to get value improvement for all possible hidden states. When an agent's state transitions and rewards do not depend on some other agent, it should not need to consider that agent's nodes as part of its hidden state. A specific compact representation along with extensions of different algorithms was proposed by Nair et al. (2005).

## Acknowledgments

We thank Martin Allen, Marek Petrik and Siddharth Srivastava for helpful discussions of this work. Marek and Siddharth, in particular, helped formalize and prove Theorem 1. The anonymous reviewers provided valuable feedback and suggestions. Support for this work was provided in part by the National Science Foundation under grants IIS-0535061 and IIS-0812149, by NASA under cooperative agreement NCC-2-1311, and by the Air Force Office of Scientific Research under grants F49620-03-1-0090 and FA9550-08-1-0181.

## Appendix A. Proof of Theorem 1

A correlation device produces a sequence of values that all the agents can observe. Let $X$ be the set of all possible infinite sequences that can be generated by a correlation device. Let $V_x(\vec{q}_0, s_0)$ be the value of the correlated joint controller with respect to some correlation sequence $x \in X$, initial nodes $\vec{q}_0$ of the agent controllers, and initial state $s_0$ of the problem. We will refer to $V_x(\vec{q}_0, s_0)$ simply as $V_x$ – the value of some sequence $x$, given the controllers for the agents. We define a *regular sequence* as a sequence that can be generated by a regular expression. Before we prove Theorem 1, we establish the following property.

**Lemma 2** *The value of any sequence, whether regular or non-regular, can be approximated within any $\epsilon$ by some other sequence.*

**Proof:** The property holds thanks to the discount factor used in infinite-horizon DEC-POMDPs. Given a sequence $x$ with value $V_x$, we can determine another sequence $x'$ such that $|V'_x - V_x| < \epsilon$. The sequence $x'$ is constructed by choosing the first $k$ elements of $x$, and then choosing an arbitrary regular or non-regular sequence for the remaining elements. As long as $k$ is chosen such that $\epsilon \geq \frac{\beta^k R_{max}}{(1-\beta)}$, then $|V'_x - V_x| < \epsilon$. $\square$

**Theorem 1** *Given an initial state and a correlated joint controller, there always exists some finite-size joint controller without a correlation device that produces at least the same value for the initial state.*

**Proof:** Let $E$ represent the expected value of the joint controller with the correlation device. Let $\mathcal{V} = \{V_x \mid x \in X\}$ be the set of values produced by all the possible correlation device sequences. Let inf and sup represent the infimum and supremum of $\mathcal{V}$ respectively.

We break the proof into two cases, depending on the relation of the expectation versus the supremum. We show in each case that a regular sequence can be found that produces at least the same value as $E$. Once such a regular sequence is found, then that sequence can be generated by a finite-state controller that can be embedded within each agent. Thus, a finite number of nodes can be added to the agents' controllers to provide equal or greater value, without using a correlation device.

Case (1) $\inf \leq E < \sup$

Based on Lemma 2, there is some regular sequence $x$ that can approximate the supremum within $\epsilon$. If we choose $\epsilon = \sup - E$, then $V_x \geq \sup - \epsilon = E$.

Case (2) $E = \sup$

If there is a regular sequence, $x$, for which $V_x = E$, we can choose that sequence. If no such regular sequence exists, we will show that $E \neq \sup$. We give a somewhat informal argument, but this can be more formally proven using cylinder sets as discussed by Parker (2002). We begin by first choosing some regular sequence. We can construct a neighborhood around this sequence (as described in Lemma 2) by choosing a fixed length prefix of the sequence. A prefix of length $k$ has a well-defined probability that is defined as $\sum_{q_c^0} P(q_c^0) \sum_{q_c^1} P(q_c^1 | q_c^0) \dots \sum_{q_c^{k-1}} P(q_c^{k-1} | q_c^{k-2})$ where $P(q_c^0)$ is the probability distribution of initial node of the correlation device and $P(q_c^i | q_c^{i-1})$ represents the probability of transitioning to correlation device node $q_c^i$ from node $q_c^{i-1}$. The set of sequences that possess this prefix has probability equal to that of the prefix. Because we assumed there exists some regular sequence which has value less than the supremum, we can always choose a prefix and length such that the values of the sequences in the set are less than the supremum. Because the probability of this set is nonzero and the value of these sequences is less than the supremum, then $E \neq \sup$, which is a contradiction.

Therefore, some regular sequence can be found that provides at least the same value as the expected value of the correlated joint controller. This allows some uncorrelated joint controller to produce at least the same value as a given correlated one. $\square$

## Appendix B. Proof of Lemma 1

For ease of exposition, we prove the lemma under the assumption that there is no correlation device. Including a correlation device is straightforward but unnecessarily tedious.

**Lemma 1** *Let $C$ and $D$ be correlated joint controllers, and let $\hat{C}$ and $\hat{D}$ be the results of performing exhaustive backups on $C$ and $D$, respectively. Then $\hat{C} \leq \hat{D}$ if $C \leq D$.*

**Proof:** Suppose we are given controllers $C$ and $D$, where $C \leq D$. Call the sets of joint nodes for these controllers $\vec{Q}$ and $\vec{R}$, respectively. It follows that there exists a function $f_i : Q_i \to \Delta R_i$ for each agent $i$ such that for all $s \in S$ and $\vec{q} \in \vec{Q}$

$$V(s, \vec{q}) \leq \sum_{\vec{r}} P(\vec{r}|\vec{q}) V(s, \vec{r}).$$

We now define functions $\hat{f}_i$ to map between the two controllers $\hat{C}$ and $\hat{D}$. For the old nodes, we define $\hat{f}_i$ to produce the same output as $f_i$. It remains to specify the results of $\hat{f}_i$ applied to the nodes added by the exhaustive backup. New nodes of $\hat{C}$ will be mapped to distributions involving only new nodes of $\hat{D}$.

To describe the mapping formally, we need to introduce some new notation. Recall that the new nodes are all deterministic. For each new node $\vec{r}$ in controller $\hat{D}$, the node's action is denoted $\vec{a}(\vec{r})$, and its transition rule is denoted $\vec{r}'(\vec{r}, \vec{o})$. Now, the mappings $\hat{f}_i$ are defined such that

$$P(\vec{r}|\vec{q}) = P(\vec{a}(\vec{r})|\vec{q}) \prod_{\vec{o}} \sum_{\vec{q}'} P(\vec{q}'|\vec{q}, \vec{a}(\vec{r}), \vec{o}) P(\vec{r}'(\vec{r}, \vec{o})|\vec{q}')$$

for all $\vec{q}$ in controller $\hat{C}$ and $\vec{r}$ in controller $\hat{D}$.

We must now show that the mapping $\hat{f}$ satisfies the inequality given in the definition of a value-preserving transformation. For the nodes that were not added by the exhaustive backup, this is straightforward. For the new nodes $\vec{q}$ of the controller $\hat{C}$, we have for all $s \in S$,

$$
\begin{aligned}
V(s, \vec{q}) &= \sum_{\vec{a}} P(\vec{a}|\vec{q}) \left[ R(s, \vec{a}) + \sum_{\vec{o}, s', \vec{q}'} P(s', \vec{o}|s, \vec{a}) P(\vec{q}'|\vec{q}, \vec{a}, \vec{o}) V(s', \vec{q}') \right] \\
&\leq \sum_{\vec{a}} P(\vec{a}|\vec{q}) \left[ R(s, \vec{a}) + \sum_{\vec{o}, s', \vec{q}'} P(s', \vec{o}|s, \vec{a}) P(\vec{q}'|\vec{q}, \vec{a}, \vec{o}) \sum_{\vec{r}'} P(\vec{r}'|\vec{q}') V(s', \vec{r}') \right] \\
&= \sum_{\vec{a}} P(\vec{a}|\vec{q}) \left[ R(s, \vec{a}) + \sum_{\vec{o}, s', \vec{q}', \vec{r}'} P(s', \vec{o}|s, \vec{a}) P(\vec{q}'|\vec{q}, \vec{a}, \vec{o}) P(\vec{r}'|\vec{q}') V(s', \vec{r}') \right] \\
&= \sum_{\vec{r}} P(\vec{r}|\vec{q}) \left[ R(s, \vec{a}(\vec{r})) + \sum_{\vec{o}, s'} P(s', \vec{o}|s, \vec{a}(\vec{r})) V(s', \vec{r}'(\vec{r}, \vec{o})) \right] \\
&= \sum_{\vec{r}} P(\vec{r}|\vec{q}) V(s, \vec{r}).
\end{aligned}
$$

$\square$